\documentclass{article}

\usepackage{PRIMEarxiv}

\usepackage[utf8]{inputenc}
\usepackage[T1]{fontenc}
\usepackage{url}
\usepackage{booktabs}
\usepackage{amsmath}
\usepackage{amssymb}
\usepackage{amsfonts}
\usepackage{nicefrac}
\usepackage{microtype}
\usepackage{fancyhdr}
\usepackage{graphicx}
\usepackage{xcolor}
\usepackage{caption}
\usepackage{graphbox}
\usepackage{subcaption}
\usepackage{float}
\usepackage{xspace}
\usepackage{algorithm}
\usepackage{algpseudocode}
\usepackage{tikz}
\usepackage{bbm}
\usepackage{multirow}
\usepackage{stmaryrd}
\usepackage{hyperref}
\usetikzlibrary{calc,patterns,
decorations.pathmorphing,
decorations.markings,positioning}

\newtheorem{theorem}{\bf Theorem}[section]
\newtheorem{definition}{Definition}[section]

\title{Equation discovery with Bayesian tree-adjoining grammars}

\author{
  Christopher A. Lindley*, Nikolaos Dervilis and Keith Worden \\
  Dynamics Research Group \\
  University of Sheffield \\
  Mappin St, Sheffield, UK \\
  \texttt{*c.a.lindley@sheffield.ac.uk} \\
}

\begin{document}
\maketitle

\begin{abstract}
Tree-Adjoining Grammars (TAGs) have recently been introduced to Nonlinear System Identification (NLSI) as a means of encoding an entire model class as a finite set of grammatical rules, from which candidate models are assembled as trees. Existing TAG-based identifiers rely on evolutionary optimisation and return point estimates of the model structure. This paper instead proposes the TAG framework within a Bayesian setting. A generative prior is defined over tree structures and their parameters, and a Reversible-Jump MCMC sampler with structure-preserving tree moves is used to infer the joint posterior over model structure, parameters and predictions. Two training objectives are considered; that is, a one-step-ahead objective with conjugate parameter proposals, and a simulation-based objective handled by likelihood-free inference. The approach is validated on a simulated polynomial NARX system, the Silverbox benchmark, and wave-loading data from the Christchurch Bay Tower, where embedding Morison's equation as a fixed initial tree yields a grey-box model that outperforms the physics-driven baseline. The results demonstrate that Bayesian TAGs are well suited to quantifying uncertainty in equation discovery for dynamical systems and to fitting physics-informed models.
\end{abstract}


\section{Introduction}

It is perhaps safe to argue that a primary goal in structural dynamics is to develop models from which predictions are consistent with past and future observations. The proof of this statement is evident in the vast and ever-growing literature surrounding this subject.

The question of interest here emerges from scrutinising what is meant by making ``better'' predictions. Concretely, model predictions can be said to be the result of a mathematical construct made to accurately represent a system of interest. Although this definition is given somewhat informally, it is enough to set the stage regarding the issue of model fidelity; that is, how well a model represents reality. What should be acknowledged about modelling is that the outcome is from an approximation of the system itself. The discrepancy between model and reality exists because several factors inherent to the system cannot be seamlessly incorporated into the corresponding model. More precisely, these factors tend to involve random effects or phenomena that are too complex to analyse and/or compute, and are often encompassed as \textit{noise} in the predictions as a result. It then becomes clear that by minimising sources of noise one can improve the fidelity of a model and, therefore, yield better predictions.

However, in practice, the solution is not nearly as straightforward as the manner in which the problem has been posed. While several means to address the issue have been proposed in all areas of science and engineering, the concern in this work is narrowed down to structural \textit{system identification} in the presence of nonlinearities. In a broader context, the term system identification is sometimes used to refer to the development of models from input and output measurements \cite{Kerschen2006}. The presence of nonlinearities can further complicate identification, given that the \textit{principle of superposition} does not generally apply to nonlinear systems \cite{Worden2001}. This caveat asserts that the functional that maps the input to the output is not clearly defined, and one is required to detect and determine the type of nonlinearity before attempting to identify the system appropriately. Such an unknown could be considered, for example, a source of \textit{epistemic} uncertainty that may hinder fidelity if identified incorrectly.

When the type of nonlinearity is known, the overall problem can then be broken down into two essential components: \textit{structure detection}\footnote{The term ``structure'' here refers to the functional form of the model formulation, and not the physical structure under investigation.}, and \textit{parameter estimation}. If unknown, however, the type of nonlinearity may be assumed, but this must be done warily. From the point of view of statistics and machine learning, these components are constituents of the \textit{model selection problem}, and since system identification is primarily a data-driven approach, both perspectives can be viewed equivalently regarding model development. 

By posing the problem in a statistical framework, one gains access to the myriad of powerful methods associated with machine learning for structure detection and parameter estimation. While no one method stands out as all-encompassing, two prevalent groups of algorithms have emerged offering the prospect of a general framework; namely, \textit{Evolutionary Optimisation} and \textit{Bayesian Inference}. Within the group of algorithms about the former, a promising development in Nonlinear System Identification (NLSI) has been the introduction of \textit{Tree-Adjoining Grammars} (TAGs) by Khandelwal, Schoukens and Toth \cite{Khandelwal2020a}. Defined originally by Joshi \textit{et al}.\ \cite{Joshi1975}, TAGs are tree-generating systems that build grammatical representations via the composition of smaller elementary trees. The original work of Khandelwal \textit{et al}.\ demonstrated that TAGs can be used in the same way to encode model classes as a set of grammatical rules. The promise of this approach for system identification was demonstrated with the development of an appropriate grammar for polynomial NARX (p-NARX) models and nonlinear state-space models, among others.

An attractive advantage of using TAGs is that they can unify many nonlinear model structures within a single framework. Moreover, the model class (or grammar) is separated from the numerical tools used for discovering system dynamics (or tree-structure representation), meaning that one is free to use any valid search algorithm as long as the rules defined by the grammar are obeyed. So far, the use of TAGs has only been explored with \textit{Evolutionary Algorithms} (EA) to automatically determine the structure and complexity of a model from data. While the EA-based approach demonstrated promising capabilities, some limitations remained that a Bayesian-based approach could overcome. One such limitation of the original method was that it produced point estimates of the model structures. A Bayesian algorithm can instead fit distributions over models in terms of both parameters and tree structure, offering a more systematic framework for model selection.

The present work aims to explore the benefits of implementing Bayesian inference as a means of determining the best model for a given system with the use of TAGs. The probabilistic nature of Bayesian inference also offers the possibility to introduce prior beliefs about the tree structure, and to quantify uncertainty in the predictions.

The outline of this paper is the following. A background on TAGs and Bayesian methods is first covered in Section \ref{sec:background}; the methodology employed is then detailed in Section \ref{sec:methodology}, whereby the procedure and application of Bayesian TAGs for dynamical systems are presented. The new approach is demonstrated using the p-NARX grammar and validated on three problems in Section \ref{sec:experiments}. The first case study considers an academic example aimed at determining whether the proposed method is capable of identifying a simulated nonlinear system. The second and third case studies involve experimental data; in particular, they look at the ``Silverbox'' nonlinear benchmark data and at wave-loading data from the Christchurch Bay Tower experiments. Finally, overall insights and suggestions are provided in Section \ref{sec:discussion}. 

\section{Background}
\label{sec:background}

\subsection{Tree-Adjoining Grammars (TAGs)}

A TAG is a type of generating grammar system used primarily to parse language of strings or sentences. Unlike other types of grammar, TAGs form a special class in which an additional set of rules is outlined to explicitly generate tree structures. In a broader sense, early definitions of a grammar in this context are from Chomsky \cite{Chomsky1956}, who defines a grammar, or a \textit{phrase-structure grammar}, as a tuple comprising the following:
\begin{enumerate}
    \item an alphabet of non-terminals $N$
    \item an alphabet of terminals $T$
    \item a distinguished start symbol $S$ in $N$
    \item a set of rules $\mathcal{P}$ that generate or replace strings in a grammar 
\end{enumerate}

This definition serves as the foundation to more elaborate structured grammars. Within the hierarchy of existing grammars, TAGs form a special class in which parsing computations, while complex, remain manageable. Another characteristic feature of TAGs that sets them apart is that they outline a set of rules that generate tree structures. The generated tree is then a visual representation of the underlying grammar that governs the language. 

\begin{figure}[ht]
    \centering
    \begin{subfigure}[c]{0.39\textwidth}
        \centering
        \includegraphics[width=\textwidth,align=c]{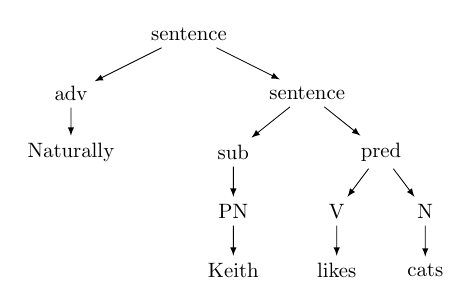}
        \caption{}
        \label{fig:nklc_tree}
    \end{subfigure}
    \hfill
    \begin{subfigure}[c]{0.59\textwidth}
        \centering
        \includegraphics[width=\textwidth,align=c]{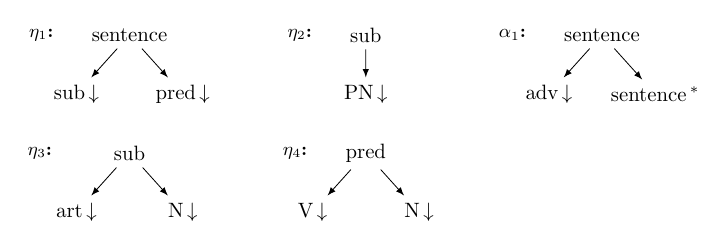}
        \caption{}
        \label{fig:nklc_tag}
    \end{subfigure}
    \caption{(a) Derived tree structure of string: ``Naturally, Keith likes cats''. (b) Tree-adjoining grammar used to parse the string. The ``sub'' in $\eta_2$ and $\eta_3$ means ``subject'' and not a substitution.}
\end{figure}

To illustrate, Figure \ref{fig:nklc_tree} shows an example of such a tree to represent the string: ``Naturally, Keith likes cats''. The key aspect of this representation is that it can be generated by defining a set of building blocks together with a set of suitable rules dictating the way in which these can be combined. For this particular example, one may define a set of smaller tree constituents, as shown in Figure \ref{fig:nklc_tag}. This set is commonly referred to as the \textit{elementary trees} of the TAG. An elementary tree can be combined with another if, and only if, its root label matches the leaf label of the other. Specifically, the combinatorial nature of TAGs is driven by two distinct operations, or rules, referred to as \textit{substitution} and \textit{adjoining}.

The set can be further partitioned to differentiate the trees governed by each operation type. The subset $I=\{\eta_1,\eta_2,\eta_3,\eta_4\}$ constitutes the \textit{initial trees}, and $A=\{\alpha_1\}$ the \textit{auxiliary trees}. The downward arrow and star symbols found in their terminal nodes indicate where substitution and adjunction take place, respectively. 

The addition of an auxiliary tree enhances the grammar and thus more elaborate sentences can be parsed. Because adjunction is carried out at non-terminal nodes in the tree, an auxiliary tree can be inserted indefinitely. The auxiliary trees and adjunction operation are thus regarded as a set of recursive replacement rules. Similarly, initial trees and the substitution operation can be regarded as a set of non-recursive replacement rules, since the allowable substitutions will eventually reach an end.

\subsection{TAG representation of dynamical systems}

To show how a dynamical system can be represented by a TAG, it is necessary to first formalise the definition of a TAG a little. The interested reader may refer to \cite{Kallmeyer2010,Joshi1997} for a more formal coverage of TAGs and their variants.

\begin{definition}[Tree-Adjoining Grammar] 
    \textit{A Tree-Adjoining Grammar is a tuple $G=\langle N,T,S,I,A \rangle$, where}
    \begin{enumerate}
        \item \textit{$N$ and $T$ are disjoint alphabets of non-terminals and terminals, respectively: $N \cap T = \emptyset$}
        \item \textit{$S \in N$ is a distinguished non-terminal symbol}
        \item \textit{$I$ and $A$ correspond to the finite set of initial trees and auxiliary trees, respectively}
    \end{enumerate}
    \textit{The set of elementary trees is thus the union $I \cup A$.}
    \label{def:tag}
\end{definition}

\begin{figure}[!h]
    \centering
    \includegraphics[width=\textwidth]{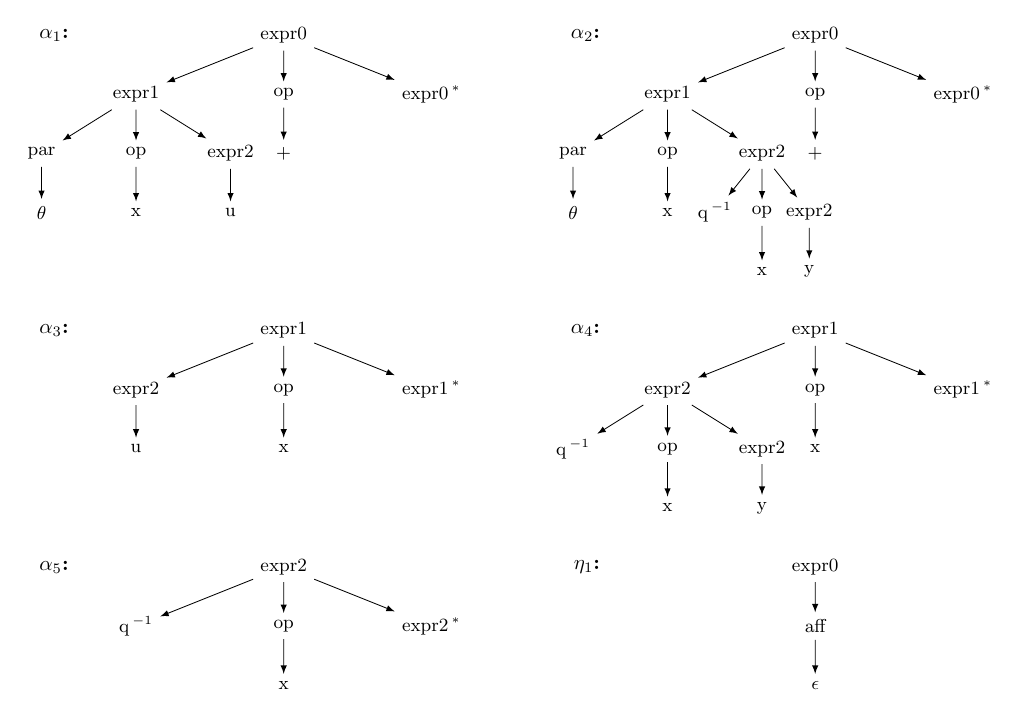}
    \caption{Initial tree $\eta_1$ and auxiliary tree set $A$ of TAG $G_N$. Symbol notation: ``expr'' - \textit{expression}, ``par'' - \textit{parameter}, ``op'' - \textit{operation}, ``$q^{-1}$'' - \textit{backward time shift}. Reproduced from \cite{Khandelwal2020a}.}
    \label{fig:tag_gn}
\end{figure}

With this definition at hand, one can now reproduce the proposal of a TAG for dynamical models from \cite{Khandelwal2020b}. The original work considers the following discrete-time input-output form of a nonlinear model,
\begin{equation}
        y_i=f\left(u_i, \ldots, u_{i-n_u}, y_{i-1}, \ldots, y_{i-n_y}, \epsilon_{i-1}, \ldots, \epsilon_{i-n_{\epsilon}}\right)+\epsilon_i
        \label{eq:discrete_siso}
\end{equation}
where $u_i,y_i \in \mathbb{R}$ denote the input-output pair at time instant $i$, $\epsilon_i \sim \mathcal{N}(0,\sigma_{\epsilon}^2)$ denotes the random noise contribution that is independent of the input, constants $n_u,n_y,n_{\epsilon}$ are integers indicating the time lags, and $f(\cdot)$ is an arbitrary nonlinear function belonging to some function space $\mathcal{M}$. Here, the function space under consideration is the set of polynomial NARX functions \cite{Billings1985}, i.e.\ with $n_{\epsilon}=0$ in the present work. This function class can approximate any continuous function arbitrarily well \cite{Stone1948}. Moreover, polynomial NARX models can be conveniently represented as discrete SISO functions \cite{Billings2013}; that is,
\begin{equation}
y_i=\sum_{k=1}^p \theta_k \prod_{j=0}^{n_u} u_{i-j}^{b_{k, j}}  \prod_{m=1}^{n_y} y_{i-m}^{a_{k, m}}+\epsilon_i,
\label{eq:poly_NARMAX_SISO}
\end{equation}
where $p$ is the number of terms, $\theta_k$ are the model parameters, $a_{k,m},b_{k,j}\in\mathbb{N}$ are the exponents for output and input terms, respectively. The current form for a polynomial NARX facilitates the TAG encoding, since the structural relationship in \eqref{eq:poly_NARMAX_SISO} is defined explicitly by a sum of weighted terms. Referring to Definition \ref{def:tag}, the corresponding TAG can now be defined as follows,

\begin{theorem}
    (Khandelwal \textit{et al}.\ \cite{Khandelwal2020b}) The TAG for a polynomial NARX model class is defined by the tuple $G_N = \langle N,T,S,I,A \rangle$ with,
    \begin{enumerate}
        \item $N= \{expr0,expr1,expr2,op,par\}$
        \item $T= \{u,y,\epsilon,+,\theta,\times,q^{-1}\}$
        \item $S=expr0$
        \item $I = \{\eta_1\}$
        \item $A = \{\alpha_1,\alpha_2,\alpha_3,\alpha_4,\alpha_5\}$
    \end{enumerate}
    where the initial tree $\eta_1$ and auxiliary trees $\{\alpha_i\}_{i=1}^5$ are depicted in Figure \ref{fig:tag_gn}. The symbol $q^{-1}$ denotes a backward time shift. It is then said that the model set $\mathcal{M}(G_N)$ is the set of all models that can be expressed as \eqref{eq:poly_NARMAX_SISO} with finite values of $p,n_u,n_y$ and $n_{\epsilon}$.
\end{theorem}

One may note from the TAG $G_N$, that the set of elementary trees can be recursively adjoined with one another, such that the constructed tree can represent any arbitrary p-NARX function of the form \eqref{eq:poly_NARMAX_SISO}. The interested reader is referred to \cite{Khandelwal2020a} for further details.

There are a few reasons why one might want to encode a model class using a TAG. The first is that the entire model class is condensed into a finite set of building blocks (i.e.\ elementary trees). This representation has the additional advantage that the grammar can be readily extended by adding new initial and auxiliary trees. Another noteworthy advantage is that the string language is generated by an underlying tree structure. This tree-based representation allows the function space to be explored by manipulating trees rather than functions directly. Such operations are often more intuitive because of their visual nature and can facilitate efficient model mixing when implementing an MCMC approach for sampling. Finally, since the search space is formulated over the set of elementary trees, the TAG remains independent of any specific model structure and can readily be combined with a wide range of search-based algorithms.

\subsection{Bayesian inference for model selection}

The highly combinatorial nature of TAGs can make equation discovery quite daunting. However, a Bayesian approach lends itself nicely to addressing this problem. From a Bayesian perspective, the model-selection problem is driven by an \textit{Occam's razor} philosophy, which proposes the idea that a model should remain simple if higher amounts of complexity are not required to explain the data. Unless favoured explicitly by the prior, Bayesian inference embodies a natural preference towards simpler models~\cite{Jefferys1992}. In particular, Occam's razor is embodied by the evidence, whereby its evaluation automatically incorporates a trade-off between model fit and model complexity~\cite{Rasmussen2000}. To illustrate this concept, Bayes' Theorem can be expressed to include the dependencies on a model (or tree structure) $\mathcal{T}_k$. That is,
\begin{equation}
    p(\Theta|\mathcal{D},\mathcal{T}_k) = \frac{ p(\mathcal{D}|\Theta,\mathcal{T}_k) p(\Theta|\mathcal{T}_k) } {p(\mathcal{D}|\mathcal{T}_k)}
    \label{eq:bayes_theorem}
\end{equation}
with the evidence given by,
\begin{equation}
    p(\mathcal{D}|\mathcal{T}_k) = \int \! p(\mathcal{D}|\Theta,\mathcal{T}_k) p(\Theta|\mathcal{T}_k) \mathrm{d}\Theta
    \label{eq:evidence}
\end{equation}
where $\mathcal{D}$ denotes the observations and $\Theta=\{\theta_1,\theta_2,\dots,\theta_p\}$ is the vector of parameters. The evidence here can be interpreted as the probability of generating the dataset after having randomly selected parameters from a given tree structure $\mathcal{T}_k$. Because the evidence is a probability distribution, simpler models are unlikely to generate the dataset. On the other hand, models that are too complex are capable of generating a richer variety of datasets besides $\mathcal{D}$, making them less likely to generate this particular dataset at random. The best model is thus the proposal that gives the highest evidence for a given set of observations, which will simultaneously happen to be the one that fits the data without added complexities~\cite{Mackay1992}.

Because the TAG $G_{N}$ can encode any p-NARX function, the resulting search space is countably infinite, which makes model identification significantly more challenging. Defining a Bayesian identifier in this setting requires two key components: (1) a joint prior over tree structures and their parameters, and (2) a set of transition kernels.

Two prevalent challenges emerge from this framework. The first is that the evidence~\eqref{eq:evidence} will likely be defined by an intractable integral, and it is thus hard (if not impossible) to solve analytically. This issue is addressed by resorting to MCMC methods to instead approximate the desired distribution. The second challenge pertains to the possible changes in dimensionality of the parameter space $\Theta_k$ for a given $\mathcal{T}_k$. This outcome would be, for example, a direct consequence of inserting an additional term-branch to the tree. While several variants of MCMC methods exist tailored to tackling this type of challenge, the focus here will be on the \textit{Reversible-Jump Markov Chain Monte Carlo} (RJ-MCMC) \cite{Green1995}, which may be thought of as a generalisation of the ubiquitous \textit{Metropolis-Hastings} (MH) algorithm \cite{Metropolis1953}. Further details on some of the concepts outlined below can be found in \cite{Bishop2006,Murphy2012,Rogers2016}.

\subsection{Reversible-Jump Markov Chain Monte Carlo}

The RJ-MCMC generalises the MH algorithm by adding the possibility to sample across subspaces of varying dimensionalities. That is, the MCMC chain can jump from a subspace $\mathcal{C}_k=\{k\}\times\mathbb{R}^k$, to another $\mathcal{C}_{k'}=\{k'\}\times\mathbb{R}^{k'}$, where the dimensionalities may differ $(k \neq k')$. The ability to jump across subspaces has powerful implications when inference is carried out over a selection of models that possess parameter sets of different sizes.

Green shows in \cite{Green1995} that \textit{detailed balance} also holds for RJ-MCMC, albeit under the \textit{dimension-matching} assumption. An explicit intuition about dimension matching is provided by considering the case in which only two subspaces exist, labelled by $k=1$ and $k=2$, and that $p(\Theta_1|k=1)$ and $p(\Theta_2|k=2)$ are proper densities in $\mathbb{R}^{m_1}$ and $\mathbb{R}^{m_2}$. To accomplish the transition from $\mathcal{C}_1$ to $\mathcal{C}_2$, it is necessary to generate an independent vector $\mathbf{u}_1\in\mathbb{R}^{n_1}$, and then set $\Theta_2$ to be some deterministic function of $\Theta_1$ and $\mathbf{u}_1$. Similarly, the same can be done backwards, by which $\Theta_1$ is defined by some function of $\Theta_2$ and $\mathbf{u}_2$, where $\mathbf{u}_2\in\mathbb{R}^{n_2}$ is generated independently of $\Theta_2$.

For dimension-matching a bijection must exist between $(\Theta_1,\mathbf{u}_1)$ and $(\Theta_2,\mathbf{u}_2)$. In particular, one requires that the lengths of $\mathbf{u}_1$ and $\mathbf{u}_2$ satisfy $m_1 + n_1 = m_2 + n_2$. The proposal distributions are then given by the distributions of $\mathbf{u}_1$ and $\mathbf{u}_2$, which are given by densities $q_1$ and $q_2$, respectively. If $j(k,\Theta_k)$ is the probability of proposing a move from the state $(k,\Theta_k)$, the general acceptance probability becomes,
\begin{equation}
    A \left ( (1,\Theta_1),(2,\Theta_2) \right ) = \min \left \{ 1,\frac{p(2,\Theta_2|\mathcal{D})j(2,\Theta_2)q_2(\mathbf{u}_{2})}{p(1,\Theta_1|\mathcal{D})j(1,\Theta_1)q_1(\mathbf{u}_{1})} \left| \frac{\partial (\Theta_2,\mathbf{u}_2)}{\partial (\Theta_1,\mathbf{u}_1)} \right| \right \}
    \label{eq:rjmcmc_acc_prob_1}
\end{equation}

There is no unique way of defining the deterministic bijection mapping $h:(\Theta_1,\mathbf{u}_1) \to (\Theta_2,\mathbf{u}_2)$. Therefore, the implementation of RJ-MCMC can be hard in practice, since one is not only required to define a suitable function, but must also evaluate the Jacobian for each move type. The reverse-move evaluation of the acceptance ratio is given by $A(z',z) = \min\{1,A(z,z')^{-1}\}$.

\subsection{RJ-MCMC for autoregressive time series}

There are two ways in which an autoregressive function can be computed, and the RJ-MCMC scheme may be adapted depending on which computation is employed. In particular, optimisation may be posed in terms of the \textit{One-Step Ahead} (OSA) error or the \textit{Model-Predicted Output} (MPO) error\footnote{These are sometimes also referred to as the \textit{prediction error} and \textit{simulation error}, respectively}. The former computes the response using measured past observations, whereas the latter estimates the response with previously estimated (or predicted) outputs over time. While either computation could be employed using the general RJ-MCMC scheme, the acceptance ratio \eqref{eq:rjmcmc_acc_prob_1} can be adapted in each case to enhance the relevance of new proposals and improve model mixing. The following subsections expand on the Bayesian inference methods proposed in each case.

\subsubsection{Approach 1: OSA-based inference}

A convenient simplification of \eqref{eq:rjmcmc_acc_prob_1} is possible when autoregressive models are linear in the parameters. This condition holds when the objective is based on the OSA error. Given the case, the proposal density can be derived to coincide with the target full conditional, as demonstrated by Troughton and Godsill in \cite{Troughton1997} for autoregressive time series. The main ideas of their approach can be illustrated by expressing equation \eqref{eq:poly_NARMAX_SISO} in the following matrix form,
\begin{equation}
    \mathbf{y}_{\ell} = \Phi_{\mathcal{T}} \Theta_{\mathcal{T}} + \boldsymbol{\epsilon}
\end{equation}
where the terms encoded by $\mathcal{T}$ are assembled columnwise into the design matrix $\Phi_{\mathcal{T}}$, the corresponding coefficients are collected in $\Theta_{\mathcal{T}}$, and the residual vector is $\boldsymbol{\epsilon}$. Since the first $\ell$ observations cannot be regressed upon, they are treated as fixed initial conditions and the response $\mathbf{y}_{\ell}$ is the original record with its first $\ell$ entries removed, of length $n_{\ell}$.

With the inclusion of additive Gaussian noise in the predictions, the likelihood can be expressed as the probability density of the observations conditioned on the parameters. That is,
\begin{equation}
    p(\mathbf{y}_{\ell}|\mathcal{T},\Theta_{\mathcal{T}},\sigma_{\epsilon}^2)
    = \mathcal{N}(\mathbf{y}_{\ell}|\Phi_{\mathcal{T}}\Theta_{\mathcal{T}},
      \sigma_{\epsilon}^2\mathbb{I}_{n_{\ell}})
    \label{eq:gaussian_likelihood}
\end{equation}

The prior over models and parameters factorises as $p(\Theta_{\mathcal{T}},\mathcal{T}) = p(\Theta_{\mathcal{T}}|\mathcal{T})\,p(\mathcal{T})$, where the parameter prior depends on the tree through its dimension. The parameter prior is defined here by the zero-mean spherical Gaussian, 
\begin{equation}
    p(\Theta_{\mathcal{T}}|\mathcal{T},\sigma_{\theta}^2)
    = \mathcal{N}\!(\Theta_{\mathcal{T}}\,|\,\mathbf{0},
      \sigma_{\theta}^{2}\mathbb{I}_{d_{\mathcal{T}}})
    \label{eq:prior_params}
\end{equation}
with $d_{\mathcal{T}} = \dim(\Theta_{\mathcal{T}})$. The posterior density is then given by,
\begin{equation}
    p(\mathcal{T},\Theta_{\mathcal{T}},\sigma_{\epsilon}^2|\mathcal{D}) \propto
    p(\mathbf{y}_{\ell}|\mathcal{T},\Theta_{\mathcal{T}},\sigma_{\epsilon}^2)\,
    p(\Theta_{\mathcal{T}}|\mathcal{T},\sigma_{\theta}^2)\,
    p(\mathcal{T})\,p(\sigma_{\epsilon}^2)
    \label{eq:posterior_parameters}
\end{equation}

An additional prior distribution on the noise variance $\sigma_{\epsilon}^2$ has been included in the formulation. Here, an inverse-gamma distribution is chosen for the noise variance prior. That is,
\begin{equation}
    p(\sigma_{\epsilon}^2) = \mathcal{IG}(\sigma_{\epsilon}^2|\alpha_{\epsilon},\beta_{\epsilon})
    \label{eq:prior_noise}
\end{equation}
which is defined with respect to the positive hyperparameters $\alpha_{\epsilon}$ and $\beta_{\epsilon}$.

The acceptance ratio \eqref{eq:rjmcmc_acc_prob_1} for a move $\mathcal{T} \to \mathcal{T}'$ can then be rewritten as follows,
\begin{equation}
    A\big((\mathcal{T},\Theta_{\mathcal{T}}),(\mathcal{T}',\Theta_{\mathcal{T}'})\big)
    = \min\left\{1,
    \frac{p(\mathcal{T}',\Theta_{\mathcal{T}'}|\dots)\,
          j(\mathcal{T}'\to\mathcal{T})\,
          q(\Theta_{\mathcal{T}}|\mathcal{T},\Theta_{\mathcal{T}'},\dots)}
         {p(\mathcal{T},\Theta_{\mathcal{T}}|\dots)\,
          j(\mathcal{T}\to\mathcal{T}')\,
          q(\Theta_{\mathcal{T}'}|\mathcal{T}',\Theta_{\mathcal{T}},\dots)}
    \right\}
    \label{eq:rjmcmc_acc_prob_3}
\end{equation}
where the Jacobian is unity because the proposal is now made directly in the new parameter subspace. For now, the noise variance $\sigma_{\epsilon}^2$, remains fixed and known. 

Since the conditional likelihood is Gaussian and linear in the parameters, and the parameter prior \eqref{eq:prior_params} is conjugate to it, the proposal distribution may be defined by a full-conditional Gaussian distribution, which can be derived analytically. Concretely,
\begin{equation}
    \begin{aligned}
        \Theta_{\mathcal{T}'} &\sim
        q(\Theta_{\mathcal{T}'}|\mathcal{T}',\Theta_{\mathcal{T}},\mathcal{D},\sigma_{\epsilon}^2)
        = p(\Theta_{\mathcal{T}'}|\mathcal{T}',\mathcal{D},\sigma_{\epsilon}^2,\sigma_{\theta}^2) \\
        &= \mathcal{N}(\Theta_{\mathcal{T}'}|\mu_{\mathcal{T}'},\Sigma_{\mathcal{T}'})
    \end{aligned}
    \label{eq:param_prop_dist}
\end{equation}
where the mean $\mu_{\mathcal{T}'}$ and covariance $\Sigma_{\mathcal{T}'}$ of the
full-conditional Gaussian are defined as,
\begin{equation}
    \begin{aligned}
        \Sigma_{\mathcal{T}'}^{-1} &= \sigma_{\epsilon}^{-2}
        \Phi_{\mathcal{T}'}^T\Phi_{\mathcal{T}'}
        + \sigma_{\theta}^{-2}\mathbb{I}_{d_{\mathcal{T}'}} \\
        \mu_{\mathcal{T}'} &= \sigma_{\epsilon}^{-2}\Sigma_{\mathcal{T}'}
        \Phi_{\mathcal{T}'}^T\mathbf{y}_{\ell}
    \end{aligned}
    \label{eq:full_conditional_moments}
\end{equation}
which correspond to the updated statistical moments of a posterior Gaussian for linear regression \cite{Bishop2006,Murphy2012}. This simplification greatly improves the effectiveness of the sampler because the proposal density coincides with the target full conditional, and the acceptance is mostly governed by the evidence for the competing model orders. The acceptance ratio can be simplified further by employing the \textit{Candidate's Identity} by Besag \cite{Besag1989}, which marginalises $\Theta_{\mathcal{T}}$ and $\Theta_{\mathcal{T}'}$ from \eqref{eq:rjmcmc_acc_prob_3}. In this case, a draw from \eqref{eq:param_prop_dist} would then need only be performed once a move has been accepted. This last step is not adopted in the present work, since the evaluation of \eqref{eq:rjmcmc_acc_prob_3} could be conducted without any apparent numerical complications, at the cost of drawing parameter vectors for rejected moves.

Finally, to address the remaining noise term, an additional Gibbs move can be included in the sampling strategy. The full conditional posterior distribution on $\sigma_{\epsilon}^2$ can also be defined analytically since the inverse-gamma prior \eqref{eq:prior_noise} is conjugate to the likelihood. The full conditional posterior is thus another inverse-gamma distribution,
\begin{equation}
    p(\sigma_{\epsilon}^2|\mathcal{T},\Theta_{\mathcal{T}},\mathcal{D})
    = \mathcal{IG}(\sigma_{\epsilon}^2|\hat{\alpha}_{\epsilon},\hat{\beta}_{\epsilon})
    \label{eq:posterior_noise}
\end{equation}
where,
\begin{equation}
    \hat{\alpha}_{\epsilon} = \alpha_{\epsilon} + \tfrac{1}{2}n_{\ell}, \qquad
    \hat{\beta}_{\epsilon} = \beta_{\epsilon}
    + \tfrac{1}{2}\left\|\mathbf{y}_{\ell} - \Phi_{\mathcal{T}}\Theta_{\mathcal{T}}\right\|^2
    \label{eq:noise_param_post}
\end{equation}
with $n_{\ell}$ being the length of the truncated response. Each proposed model move is therefore followed by a Gibbs update of the noise variance, drawn directly from \eqref{eq:posterior_noise} while $\mathcal{T}$ and $\Theta_{\mathcal{T}}$ are held at their current values. The moments in \eqref{eq:full_conditional_moments} are then recomputed with the updated $\sigma_{\epsilon}^2$ before the next model move is proposed. A successful implementation of the scheme outlined here for NLSI can also be found in \cite{Champneys2025}, in which the authors construct a Bayesian framework for the SINDy algorithm.

\subsubsection{Approach 2: MPO-based inference}

Unfortunately, when the predictions are based on the MPO setting, the parameter proposals can no longer be drawn from a full conditional as they can for OSA predictions. A free-run simulation is a nonlinear function of the whole parameter vector, so no conjugacy is available, and the simulation itself is prone to diverge.

\textit{Approximate Bayesian Computation} (ABC) \cite{Tavare1997} offers a way to address this issue. ABC is a rejection-type algorithm designed to overcome complications regarding the likelihood evaluation\footnote{This type of algorithm is sometimes referred to as \textit{likelihood-free inference}, or \textit{simulation-based inference}.}. In addition to circumventing the need to evaluate the evidence, ABC also removes the need to evaluate the likelihood, at the cost of targeting an approximation to the true posterior. Likelihood-free techniques are often used when the likelihood is intractable or unknown, but simulations can be carried out with relative ease. In the present case, likelihood-free techniques also lend themselves to dealing with divergent simulations, since models that produce large discrepancies are rejected during inference.

While the current setting, i.e.\ encoding a NARX model class, and minimising the residual error, could in principle be handled with an explicitly defined likelihood, an ABC approach is adopted here because it generalises more readily to more complex cases. Two considerations motivate this choice:
\begin{enumerate}
    \item The grammar $G_N$ can be extended to include \textit{Moving-Average} (MA) auxiliary trees, taking the model class from NARX to NARMAX. A likelihood-free scheme adapts to this change without any modification to its implementation, whereas a likelihood-based treatment would have to be re-derived for such an extension.
    \item Large datasets may make pointwise comparison prohibitively expensive. In such cases, one may wish to pre-process the data into a set of key dynamical features, and match those features rather than the pointwise error. A likelihood-based approach would then require a sampling distribution over these summary statistics, which ABC avoids. The implementation would simply involve defining the discrepancy on these features.
\end{enumerate}

The ABC approximation to the joint posterior over $\mathcal{T}$ and the corresponding parameter vector $\Theta_{\mathcal{T}}$ is,
\begin{equation}
  p(\Theta_{\mathcal{T}},\mathcal{T}|\mathcal{D})
  \;\approx\;
  p_{\varepsilon}(\Theta_{\mathcal{T}},\mathcal{T}|\mathcal{D})
  \;\propto\;
  p(\mathcal{T})\,p(\Theta_\mathcal{T}\mid\mathcal{T})
  \int p\!\left(\mathbf{y}^*\mid\Theta_{\mathcal{T}},\mathcal{T},\mathcal{D}\right)
  \mathbbm{1}_{\{\rho(\mathbf{y}^*,\mathbf{y})<\varepsilon\}}\
  \mathrm{d}\mathbf{y}^* ,
  \label{eq:abc_posterior}
\end{equation}
where $p(\mathbf{y}^*\mid\Theta_{\mathcal{T}},\mathcal{T},\mathcal{D})$ is the distribution of data simulated from the candidate model. The likelihood is thereby replaced by the probability that a simulation from $(\mathcal{T},\Theta_{\mathcal{T}})$ falls within the tolerance around $\mathbf{y}$. In practice, the integral in equation \eqref{eq:abc_posterior} is never explicitly evaluated. Instead, the chain is run over the augmented space, with target,
\begin{equation}
  p_{\varepsilon}(\mathcal{T},\Theta_{\mathcal{T}},\mathbf{y}^*\mid\mathcal{D})
  \;\propto\;
  p(\mathcal{T})\,p(\Theta_{\mathcal{T}}\mid\mathcal{T})\,
  p\!\left(\mathbf{y}^*\mid\mathcal{D},\Theta_{\mathcal{T}},\mathcal{T}\right)
  \mathbbm{1}_{\{\rho(\mathbf{y}^*,\mathbf{y})<\varepsilon\}}
  \label{eq:abc_augmented}
\end{equation}
A move proposes $(\mathcal{T},\Theta_{\mathcal{T}})$ and then simulates $\mathbf{y}^{*}$ from it, so the simulator density appears both in the target and in the proposal, thereby cancelling from the acceptance ratio and leaving only the priors and the tolerance indicators. Therefore, the acceptance probability is given by,
\begin{equation}
    A\big((\mathcal{T},\Theta_{\mathcal{T}}),(\mathcal{T}',\Theta_{\mathcal{T}'})\big)
    = \min\left\{1,
    \frac{p(\mathcal{T}',\Theta_{\mathcal{T}'})\,
          j(\mathcal{T}'\to\mathcal{T})\,
          q_{\mathcal{T}'\to\mathcal{T}}(\mathbf{u}')}
         {p(\mathcal{T},\Theta_{\mathcal{T}})\,
          j(\mathcal{T}\to\mathcal{T}')\,
          q_{\mathcal{T}\to\mathcal{T}'}(\mathbf{u})}
    \right\} \;\mathbbm{1}_{\{\rho(\mathbf{y}^*,\mathbf{y})<\varepsilon\}}
    \label{eq:acceptance_ratio}
\end{equation}
where $\mathbf{u}$ and $\mathbf{u}'$ are the auxiliary vectors of the forward and reverse moves, drawn from the proposal densities $q_{\mathcal{T}\to\mathcal{T}'}$ and $q_{\mathcal{T}'\to\mathcal{T}}$, respectively, with $\mathbf{u}\in\mathbb{R}^{n_{\mathcal{T}\to\mathcal{T}'}}$ and $\mathbf{u}'\in\mathbb{R}^{n_{\mathcal{T}'\to\mathcal{T}}}$. For the jumps employed here, a (birth) move sets the mapping $\Theta_{\mathcal{T}'}=[\Theta_{\mathcal{T}};\,\mathbf{u}]$, with $\mathbf{u}\sim q_{\mathcal{T}\to\mathcal{T}'}(\cdot)$, taken here to be a zero-mean Gaussian with standard deviation $\sigma_u$. In the reverse (death) direction, no auxiliary variables are required, so $n_{\mathcal{T}'\to\mathcal{T}}=0$, $q_{\mathcal{T}'\to\mathcal{T}}\equiv1$, and the Jacobian determinant is, once again, unity. The proposed state is thus accepted with probability given by the $\min\{\cdot\}$ term, but only if the simulated output satisfies $\rho(\mathbf{y}^*,\mathbf{y})<\varepsilon$. Otherwise, it is rejected, and the chain remains at $(\mathcal{T},\Theta_{\mathcal{T}})$.

The downside of this approach is that the acceptance rate tends to be low. This happens when the posterior diverges away from the prior, so that few prior draws land in regions covered by the posterior mass. Furthermore, $\rho(\cdot,\cdot)$ and $\varepsilon$ are additional hyperparameters that must be tuned carefully. Concretely, a tolerance that is too large may yield poor posterior approximations, and a tolerance that is too small may prevent a chain from moving at all.

The acceptance rate can be improved by introducing a Markov kernel $q(\Theta^*\mid\Theta)$, so that a new parameter vector is proposed from the current one rather than drawing them at random from the prior. Marjoram \textit{et al}.\ \cite{Marjoram2003} showed that the resulting ABC-MCMC algorithm returns an invariant-density approximation to the target $p_{\varepsilon}(\mathcal{T},\Theta_{\mathcal{T}},\mathbf{y}^*\mid\mathcal{D})$, and it is thus a valid adaptation of the standard ABC rejection sampler.

Furthermore, the method proposed by \cite{Baragatti2012} suggests refining the ABC-MCMC sampler with a parallel tempering scheme to improve mixing and mitigate the chances of a chain becoming trapped in a local mode. The idea is to run $C$ chains in parallel, each targeting a tempered version of the target distribution. Tempered distributions become flatter at higher ``temperatures'', allowing their chain to traverse the space between modes more easily. At intervals, a swap of the states of two chains is proposed and accepted or rejected by a Metropolis rule, so that states discovered by the exploratory chains can propagate down to the chain that targets the distribution of interest.

Tempering is enforced here not by explicitly flattening the target, but via the tolerance. In particular, a sequence of levels $\boldsymbol{\varepsilon}=(\varepsilon_0,\varepsilon_1,\dots,\varepsilon_{C-1})$, is defined with $\varepsilon_0<\varepsilon_1<\dots<\varepsilon_{C-1}$, and the $i$-th chain targets
$p_{\varepsilon_i}(\mathcal{T},\Theta_{\mathcal{T}},\mathbf{y}^*\mid\mathcal{D})$. Loosening the tolerance widens the region of
non-zero pseudo-likelihood, which raises the acceptance rate of the larger jumps proposed
by the higher-order chains. Under the uniform ABC kernel of equation~\eqref{eq:abc_augmented}, and with a common prior across chains, a swap between chains $i$ and $j>i$ is accepted whenever the state held by the looser chain $j$ also satisfies the tighter tolerance $\varepsilon_i$. Inference is then based on the ``coldest'' chain $0$, which targets the closest approximation to the true posterior.

\subsection{Prior distribution, $p(\Theta_{\mathcal{T}},\mathcal{T})$}
\label{sec:prior_distribution}

So far, the prior has only been partly defined. The full definition is provided in this section, and it applies to both MCMC approaches covered above. To begin, the model prior component $p(\mathcal{T})$ is defined through the probabilities assigned to each event in the generative process that produces a given tree structure, following the approaches of~\cite{Jin2020,Brence2021}. Specifically, the generative process assigns probabilities to the three defining characteristics of a discrete p-NARX function: the number of terms, the term degree, and the time-lag offsets. These characteristics relate directly to the auxiliary trees of the TAG; that is, a new term is introduced by $\alpha_1$ or $\alpha_2$, a new factor by $\alpha_3$ or $\alpha_4$, and a lag offset by $\alpha_5$.

Consider an arbitrary tree $\mathcal{T}$. Let a branch $\mathcal{T}_{\mathrm{expr0}} \in \mathcal{T}$, rooted at a node labelled ``expr0'', encode a single term of a p-NARX expression. This branch is itself composed of an adjoined collection of $\mathcal{T}_{\mathrm{expr1}}$ subbranches, each encoding one factor of the term, and each $\mathcal{T}_{\mathrm{expr1}}$ subbranch is in turn composed of a collection of $\alpha_5$ auxiliary trees that together encode the lags of that factor.

The branch $\mathcal{T}_{\mathrm{expr0}}$ is generated bottom-up. For each factor, the number of $\alpha_5$ trees is first sampled at random, and these are then jointly adjoined to an $\alpha_{\{3,4\}}$ tree to form the corresponding $\mathcal{T}_{\mathrm{expr1}}$ subbranch. This process is repeated once for each factor in the term, and the resulting collection of $\mathcal{T}_{\mathrm{expr1}}$ subbranches is adjoined together to an $\alpha_{\{1,2\}}$ tree to complete $\mathcal{T}_{\mathrm{expr0}}$.

Formally, the generative process is defined as follows, where $d$ denotes the term degree (the number of factors in the term) and $n_l$ the number of lags in a given factor, drawn respectively from a Poisson rate $\lambda_d$ and a Geometric continuation probability $\lambda_l \in (0,1)$:
\begin{equation}
\label{eq:term_construct}
\begin{aligned}
    n_l &\sim \text{Geometric}(\lambda_l)
        &\qquad (d-1) &\sim \text{Poisson}(\lambda_{d}) \\
    \alpha_{\{3,4\}} &\sim \text{Uniform}\{\alpha_3,\alpha_4\}
        &\qquad \alpha_{\{1,2\}} &\sim \text{Uniform}\{\alpha_1,\alpha_2\} \\
    \mathcal{T}_{\mathrm{expr2}} &= \alpha_5^{(n_l)}\left\llbracket\mathrm{expr2},\alpha_5^{(n_l-1)}\llbracket\mathrm{expr2}, \dots \rrbracket \right\rrbracket
        &\qquad \mathcal{T}_{\mathrm{expr1}} &= \alpha_{\{3,4\}} \llbracket \mathrm{expr2},\mathcal{T}_{\mathrm{expr2}}\rrbracket \\
    \mathcal{T}_{\mathrm{expr0}} &= \alpha_{\{1,2\}}\left\llbracket\mathrm{expr1},\mathcal{T}_{\mathrm{expr1}}^{(d-1)}\llbracket\mathrm{expr1}, \mathcal{T}_{\mathrm{expr1}}^{(d-1)}\llbracket\mathrm{expr1}, \dots \rrbracket \rrbracket \right\rrbracket
\end{aligned}
\end{equation}
where adjoining $\psi'$ into $\psi$ at node $\nu$ is denoted by $\psi\llbracket\nu,\psi'\rrbracket$, and $\alpha_{\{i,j\}} \sim \text{Uniform}\{\alpha_i,\alpha_j\}$ denotes a uniform random choice between the two auxiliary trees $\alpha_i$ and $\alpha_j$, each selected with probability $1/2$. The probability assigned to $\mathcal{T}_{\mathrm{expr0}}$ is defined as,
\begin{equation}
    p(\mathcal{T}_{\mathrm{expr0}}) = \left (\frac{1}{2} \right )^{d} \times \frac{\lambda_{d}^{d-1}e^{-\lambda_{d}}}{(d-1)!} \times \frac{d!}{\prod c_k!} \times \prod_{d} (1-\lambda_l)\lambda_l^{n_{l}^{(d)}}
    \label{eq:distr_terms}
\end{equation}

Equation \eqref{eq:distr_terms} is the product of the probabilities assigned at each stage of the generative process, under the assumption that the factors encoded by each $\mathcal{T}_{\mathrm{expr1}}$ are drawn independently. The multinomial term in the equation corrects for the fact that a term with repeated factors can be encoded by multiple orderings of the adjunction process, as defined in~\eqref{eq:term_construct}. The variable $c_k$ is then the multiplicity of each distinct factor type among $d$ factors.

For the entire tree structure $\mathcal{T}$, the prior distribution is given by the combined contribution of every $\mathcal{T}_{\mathrm{expr0}}$ branch in $\mathcal{T}$. Under the assumption that the terms are drawn independently, the result yields a Poisson point process of the form,
\begin{equation}
    p(\mathcal{T}) = \lambda_{t}^{n_t}e^{-\lambda_t}\prod_{\mathcal{T}_{\mathrm{expr0}} \in \mathcal{T}} p(\mathcal{T}_{\mathrm{expr0}})
    \label{eq:prior_model}
\end{equation}
where $\lambda_{t}$ denotes the expected rate of terms, $n_t$ the number of terms in the model, and $p(\mathcal{T}_{\mathrm{expr0}})$ the distribution over individual terms defined in \eqref{eq:distr_terms}. Note that the tree is constructed such that no repeated terms are possible, and a correction factor is thus not necessary in this formulation. There is, however, an additional step in the generative process that is necessary to ensure the desired models are sampled more often. The idea is to help promote pure-power terms, such as squared or cubed terms, as these are more likely to be found in equations derived from engineering systems. This bias is introduced here by reformulating the probability assigned to a given term as a mixture,
\begin{equation}
    p(\mathcal{T}_{\mathrm{expr0}}) = (1-\lambda_{m})q_{c}(\mathcal{T}_{\mathrm{expr0}}) + \lambda_{m}q_{p}(\mathcal{T}_{\mathrm{expr0}})
    \label{eq:prior_mix_term}
\end{equation}
where $q_{c}$ corresponds to the original equation~\eqref{eq:distr_terms} and $q_{p}$ is an additional component responsible for assigning probabilities to pure-power terms, defined as,
\begin{equation}
    q_{p}(\mathcal{T}_{\mathrm{expr0}}) = \left (\frac{1}{2} \right ) \times \frac{\lambda_{p}^{d-1}e^{-\lambda_{p}}}{(d-1)!} \times (1-\lambda_l)\lambda_l^{n_{l}}
    \label{eq:power_distr_terms}
\end{equation}

The distribution is now a mixture of these probability measures, which are weighted by the mixing parameter $\lambda_{m}$. The generative process now chooses to either produce a subbranch for some interaction term, with probability $(1-\lambda_{m})$, or one for a pure-power term, with probability $\lambda_{m}$. If the latter is picked, then the multinomial term returns unity and the factor only needs to be sampled once. In the original formulation, a pure-power term would have been drawn by randomly sampling the same factor $d$ times in a row, making them less likely to emerge. Simultaneously, pure-power terms are preferred over interaction terms by the prior, since the former draws weight from both terms in equation~\eqref{eq:prior_mix_term}, while the latter can only be sampled from $q_{c}(\mathcal{T}_{\mathrm{expr0}})$.

Overall, the prior on tree structure serves to regularise the complexity of the encoded model. The model size, for example, is regularised by adjusting the expected number of terms $\lambda_{t}$. Similarly, the term degree and the number of lags in a factor are regularised by $\lambda_{d}$ and $\lambda_l$, respectively. These quantities are hyperparameters that must be specified in advance.

Regarding the parameters associated with a given tree structure, the prior distribution is given by the zero-mean spherical Gaussian distribution defined in~\eqref{eq:prior_params}. This Gaussian prior on the parameters was alluded to earlier in deriving the full conditional Gaussian as the proposal distribution. The same prior is also used for the MPO-based inference approach. A (hyper)prior distribution on $\sigma_{\theta}^2$ could also be defined in the same manner as it was done for the noise variance in Equation~\eqref{eq:prior_noise}. However, this value was left as a fixed hyperparameter.

\subsection{Transition kernels, $j(\mathcal{T} \to \mathcal{T}')$}

A set of transition operations are required to traverse the model space. The transition from one tree to another is conducted by explicitly altering the tree structure. Such transformations are carefully defined to ensure the model space is explored efficiently. A tree alteration may, for example, involve inserting an additional branch to the current structure such that a new term is introduced to the overall expression. Conversely, a branch could be pruned away, thus removing a term from the expression. By recursively employing these alterations, a sequence of tree structures is proposed and evaluated to determine the ones that give the highest evidence given the observations.

Since the modifications are explicitly exercised on the tree structure, one can draw operators from other tree-based models that have already been explored in the literature. For example, Jin \textit{et al}.\ \cite{Jin2020} defined seven different move type operators to transition between structures of linearly-mixed tree representations for \textit{Bayesian Linear Regression}. Another renowned example is that from Chipman \textit{et al}.\ \cite{Chipman1998,Chipman2010}, in which the generation of a new tree is conducted by employing one of four different moves; namely, \textit{growing, pruning, changing} and \textit{swapping}. These move-type proposals were demonstrated to work successfully for both their \textit{Classification And Regression Tree} (CART) and \textit{Bayesian Additive Regression Tree} (BART) models. A slight variation of the latter can be found in \cite{Zhang2020} for \textit{Treed Gaussian Processes} (TGPs) \cite{Gramacy2008}, in which a \textit{rotation} move-type is shown to be better suited than \textit{swapping} for manipulating TGPs.

These examples are a few of several tree-based models used in learning problems. It becomes clear that the move-types are application dependent. Inspired by the work above, the following move types are proposed here:

\begin{enumerate}
    \item \textbf{Add term:} A random subbranch, $\mathcal{T}_{\mathrm{expr0}}$, is generated
    via the process defined in \eqref{eq:term_construct} and adjoined to the root
    node labelled ``expr0''. If the generated term is already present, the proposal is
    discarded. This step enforces the absence of repeated terms assumed
    in~\eqref{eq:prior_model}.

    \item \textbf{Remove term:} A branch $\mathcal{T}_{\mathrm{expr0}}$ is
    chosen uniformly at random. All of its descendants except those in the
    rightmost subtree are discarded, and the chosen node is then replaced by the
    remaining rightmost subtree.

    \item \textbf{Add factor:} A branch $\mathcal{T}_{\mathrm{expr0}}$ is
    chosen uniformly at random. A subbranch $\mathcal{T}_{\mathrm{expr1}}$ is then selected
    with probability proportional to the prior weight of the resulting term, as
    given in \eqref{eq:prior_factor_weight}, and adjoined to an ``expr1'' node
    within $\mathcal{T}_{\mathrm{expr0}}$.

    \item \textbf{Remove factor:} A branch $\mathcal{T}_{\mathrm{expr0}}$
    with at least two factors is chosen uniformly at random. One of its
    distinct factor subbranches is then selected with probability
    proportional to the prior weight of the reduced term. The corresponding node is pruned while
    retaining the rightmost subtree, which then replaces the chosen node.

    \item \textbf{Add lag:} A lag-type auxiliary tree $\alpha_5$ is picked at
    random from $G_{N}$. A nonterminal node labelled ``expr2'' is then chosen
    uniformly from the current tree, and the auxiliary tree is adjoined to that
    node.

    \item \textbf{Remove lag:} A nonterminal node labelled ``expr2'' that carries
    an $\alpha_5$ adjunction is chosen uniformly from the current tree, excluding
    any node whose reduction would yield $y_i$. The leftmost side of the selected node
    corresponds to a lag-type auxiliary tree, which is pruned away while retaining
    the rightmost subtree. The chosen node is then replaced by the remaining
    rightmost subtree.

    \item \textbf{Swap input:} A subtree with yield $u_{i-1}$ is
    chosen uniformly from the current tree, and is then replaced by a subtree with
    yield $y_{i-1}$.

    \item \textbf{Swap output:} A subtree with yield $y_{i-1}$ is chosen uniformly
    from the current tree. It is then replaced by a subtree with yield $u_{i-1}$.

    \item \textbf{Stay:} A new set of parameters is proposed within the current
    state.
\end{enumerate}

It should be highlighted that the ``add factor'' proposals is a little more subtle than uniformly choosing the relevant auxiliary tree from the TAG. The generative process here is devised to have the prior and proposal distributions cancel in the acceptance ratio. Therefore, a new factor must be drawn with probability propotional to the prior weight of the resulting term. Concretely, when the ``add factor'' move is employed, a new tree subbranch is drawn with probability,
\begin{equation}
    p(\mathcal{T}_{\mathrm{expr1}}) = \frac{p(\mathcal{T}_{\mathrm{expr0}} \llbracket \mathrm{expr1}, \mathcal{T}_{\mathrm{expr1}} \rrbracket )}
    {\sum_{\mathcal{T}_{\mathrm{expr1}'} \in \mathcal{F}} p(\mathcal{T}_{\mathrm{expr0}} \llbracket \mathrm{expr1}, \mathcal{T}_{\mathrm{expr1}'} \rrbracket )}
    \label{eq:prior_factor_weight}
\end{equation}
where $\mathcal{F}$ denotes the space of tree structures encoding all possible factors with arbitrary time lags. The new problem here is that the current generative process could propose a factor with any arbitrary number of lags, prior to being subsequently adjoined to some subbranch $\mathcal{T}_{\mathrm{expr0}}$ in the tree. This outcome means dealing with a support of factors that is countably infinite, which complicates the way in which equation \eqref{eq:prior_factor_weight} can be evaluated. To address this issue, the lag support is capped, thus making $\mathcal{F}$ finite. The nomarlising constant in \eqref{eq:prior_factor_weight} can then be carried out over a finite sum of $|\mathcal{F}|$ evaluations of $p(\mathcal{T}_{\mathrm{expr0}} \llbracket \mathrm{expr1}, \mathcal{T}_{\mathrm{expr1}} \rrbracket$).

\begin{figure}[!h]
    \centering
    \includegraphics[width=\textwidth]{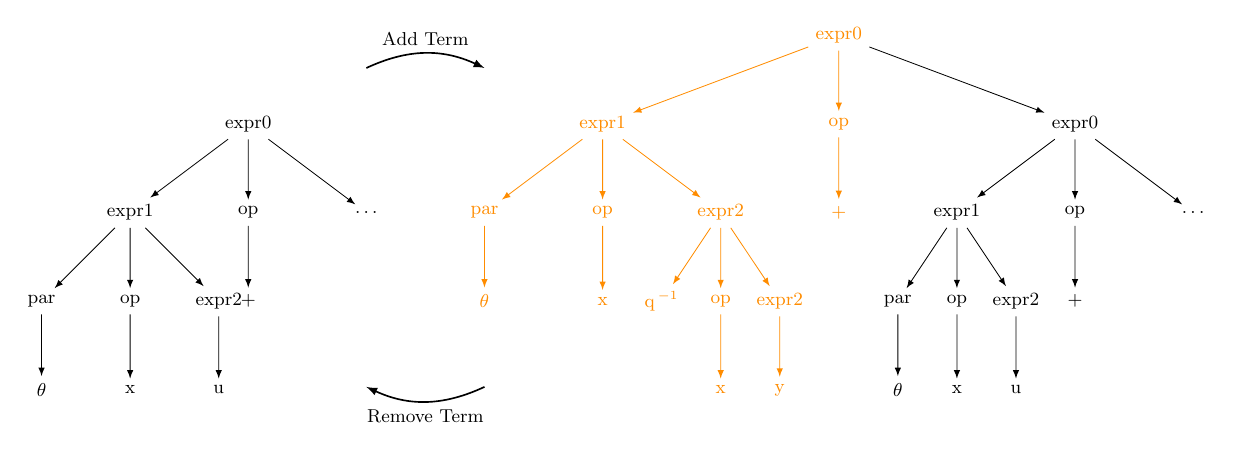}
    \caption{Illustration of the add/remove term move type.}
    \label{fig:add_remove_term}
\end{figure}

One may note that all moves are reversible, so that detailed balance is preserved. These moves are expressed entirely through adjunction, ensuring the proposed structure remains valid with respect to the TAG. Note, however, that ``add term'' and ``add factor'' adjoin an auxiliary tree generated from the countably infinite family implied by \eqref{eq:term_construct}, rather than one selected from the finite set $G_{N}$. It is this distinction that necessitates the prior-proportional selection described above. The first three sets of moves (add/remove term, add/remove factor, and add/remove lag) ensure that a valid model is sampled, while the fourth set (swap input/output) has been included to enhance model mixing. To illustrate, Figure \ref{fig:add_remove_term} shows the simplest case of the add/remove term jump, as an example.

\section{Methodology}
\label{sec:methodology}

The methodology proposed here makes use of all the concepts outlined above. A point worth highlighting at this stage is that one must choose several hyperparameters and address a few practical considerations when adopting this approach. These considerations have a strong influence on the effectiveness of the proposed algorithm. Therefore, it is important to acknowledge that the choices made in the present work may not be optimal, and that one must adapt them to the problem at hand.

\subsection{MCMC adaptation schemes}
\label{sec:mcmc_adaptation}

Additional measures were required to guarantee the success of the MPO-based sampler. Concretely, the step size of the parameter proposal $s_c$, and the sequence of tolerance levels $(\varepsilon_0,\dots,\varepsilon_{C-1})$ were made to adapt with respect to the data. An additional annealing phase was thus added prior to running the identifier, during which these variables were adjusted. The step size was adapted via the Robbins-Monro stochastic approximation method \cite{Garthwaite2016}. Each parallel chain carried its own width and these were updated only on the stay move. The adaptation was tuned to aim for a target acceptance rate of $0.234$ across chains.

As for the tolerance ladder, it is suggested in \cite{Baragatti2012} that the sequence of tolerances should be chosen so that they are evenly spaced on a logarithmic scale, or that their inverses are evenly spaced geometrically. The authors also suggest having an adaptive algorithm to adjust the sequence of tolerance levels. Therefore, the approach adopted here employs an adaptive scheme that targets the cold tolerance $\varepsilon_0$, followed by adjusting the remaining tolerances such that their inverses are geometrically spaced. This adaptation schedule is in the spirit of \cite{DelMoral2012}, whereby a \textit{Sequential Monte Carlo} approach is embedded in the ABC algorithm.

The way the adaptation works is by varying $\varepsilon_0$ in stages during the annealing phase. The schedule begins with a ``loose'' tolerance, and at the end of each stage, it is updated with respect to the smallest distance observed so far. The cold tolerance is tightened onto a quantile. In this case, the tolerance was brought to the 20th percentile of the distances visited by the cold chain. The whole ladder could then be adjusted geometrically above the new floor. Once completed, both the step sizes and the tolerances were frozen, and the MCMC sampler was initiated as normal.

\subsection{Hyperparameter selection}
\label{sec:hyperparameter_selection}

In addition to having established suitable prior distributions and valid move-type proposals, several hyperparameters of the algorithm remained that had to be selected upfront. Among such hyperparameters were those found in the prior distributions. Other hyperparameters belonged to the RJ-MCMC sampler, such as the number of samples and transition probabilities. Unless stated otherwise, the hyperparameters used in the following experiments are the ones listed in Tables \ref{tab:hyps_priors} and \ref{tab:hyps_rjmcmc}.
\begin{table}[tb]
\caption{Hyperparameters of the chosen prior distributions.}
\label{tab:hyps_priors}
\centering
\begin{tabular}{lcc}
\hline
Hyperparameter & Notation & Value \\ \hline
\hline
\multicolumn{3}{l}{\textit{Structure prior (both methods)}} \\ \hline
Term rate (prior mean number of terms) & $\lambda_t$      & $\{1,\,5,\,20\}$ \\
Term degree rate                       & $\lambda_d$      & $0.3$ \\
Lag decay                              & $\lambda_l$      & $0.6$ \\
Pure-power mixture weight              & $\lambda_m$      & $0.5$ \\
Pure-power degree rate                 & $\lambda_p$      & $0.7$ \\[3pt] \hline
\multicolumn{3}{l}{\textit{Parameter prior (both methods)}} \\ \hline
Parameter prior std.\ deviation        & $\sigma_{\theta}$   & $1.0$ \\[3pt] \hline
\multicolumn{3}{l}{\textit{Noise prior (OSA only)}} \\ \hline
Noise prior shape                      & $\alpha_{\epsilon}$ & $1\times10^{-6}$ \\
Noise prior scale                      & $\beta_{\epsilon}$  & $1.0$ \\ \hline
\end{tabular}
\end{table}
\begin{table}[tb]
\caption{Default hyperparameters defined for the RJ-MCMC samplers.}
\label{tab:hyps_rjmcmc}
\centering
\begin{tabular}{lcc}
\hline
Hyperparameter & Notation & Value \\ \hline
\hline
\multicolumn{3}{l}{\textit{Shared by both methods}} \\ \hline
Number of samples (OSA / MPO)  & $n_s$            & $30{,}000$ / $60{,}000$ \\
Burn-in samples (OSA / MPO)    & $n_{\mathrm{burn}}$       & $0.5n_s$ / $0.75n_s$ \\
Thinning interval              & $n_{\mathrm{thin}}$       & $5$ \\
Lag window & $\ell$      & $10$ \\
Stay rate                      & $p_o$            & $0.30$ \\
Add/remove term rate           & $p_{at}/p_{rt}$  & $0.12/0.12$ \\
Add/remove factor rate         & $p_{af}/p_{rf}$  & $0.08/0.08$ \\
Add/remove lag rate            & $p_{al}/p_{rl}$  & $0.08/0.08$ \\
Input/output swap rate         & $p_{is}/p_{os}$  & $0.07/0.07$ \\[3pt] \hline
\multicolumn{3}{l}{\textit{MPO only (tolerance ladder)}} \\ \hline
Number of steps                & $C$              & $12$ \\
Swap interval                  & $n_{\mathrm{swap}}$       & $25$ \\
Cold step margin               & ---              & $1.02\,d_{\min}$ \\[3pt] \hline
\multicolumn{3}{l}{\textit{MPO only (annealing)}} \\ \hline
Annealing stages               & $S$              & $5$ \\
Annealing share of budget      & ---              & $0.5n_s$ \\
Tightening quantile            & $q$              & $0.2$ \\[3pt] \hline
\multicolumn{3}{l}{\textit{MPO only (proposals and calibration)}} \\ \hline
Auxiliary proposal std.\ deviation & $\sigma_u$   & $0.1$ \\
Initial random-walk width      & $s_0$            & $0.1$ \\ \hline
\end{tabular}
\end{table}

The term rate in Table \ref{tab:hyps_priors} presents three values $\lambda_t = \{1,\,5,\,20\}$. These values have been chosen to promote the exploration of models with increasing degrees of complexity, under the assumption that more terms lead to models capable of explaining a wider range of observations. The experiments were thus repeated three times with $\lambda_t$ set to each of these values. Concretely, $\lambda_t = 1$ for simpler model structures, $\lambda_t = 5$ for models with moderate complexity, and $\lambda_t = 20$ for models with a significantly higher number of terms. As noted in the table, this configuration applies to both OSA and MPO approaches.

Another point worth highlighting is on the number of samples assigned to each approach. The reason why the $n_s$ assigned to the MPO approach doubles that of the OSA is to account for the added annealing phase. Overall, the current configuration was devised to yield the same number of post burn-in samples for either approach, and thus aid in making fair comparisons between their results. In addition to these choices, all runs of the algorithm were carried out over four parallel chains, each initiated with a different random seed.

\subsection{Validation criteria}

Whenever possible, the available (or generated) data were partitioned into three independent subsets: a training set $\mathcal{D}_{\mathrm{train}}$, a validation set $\mathcal{D}_{\mathrm{val}}$, and a test set $\mathcal{D}_{\mathrm{test}}$. The model structure and parameters were estimated exclusively using the training set, while the independent validation set was used to assess predictive performance and guide model selection. After selecting the configuration that achieved the best performance on the validation set, the final generalisation error was evaluated using predictions on the independent test set.

The goodness of fit was calculated here using the \textit{Normalised Mean-Square Error} (NMSE), which is defined by,
\begin{equation}
    \mathrm{NMSE} = \frac{100}{N\sigma_{y}^2} \sum_{i=1}^N (y_i - y_i^*)^2
    \label{eq:nmse}
\end{equation}
where the star denotes the estimated quantity. It should be emphasised that the NMSE is always computed with MPO predictions, irrespective of the training objective used.  

\subsection{Predictions via MAP and forward integration}

A quick estimate of the response could be computed using the \textit{Maximum-A-Posteriori} (MAP) estimates of both the model posterior density and parameter posterior densities. This approach simply involves evaluating the most frequent model structure, parameterised by the MAP values of its corresponding parameter posterior. Another more elaborate approach involves propagating the estimated uncertainties forward towards the response predictions. The process for generating these simulations was conducted by following three steps:
\begin{enumerate}
    \item Sample a model structure from the estimated model posterior.
    \item Sample a set of parameters, conditioned on the model structure, from the estimated parameter posterior.
    \item Simulate a sequence of response predictions with respect to the sampled model structure and associated parameters.
\end{enumerate}
By repeating these steps sufficiently many times, a Monte Carlo approximation of the prediction mean and variance was obtained from the time-series realisations. Step Three, however, was slightly more involved. As it stands, this propagation process underestimates the prediction uncertainties. To get a better estimate, an additional random draw was made explicitly from the predictive distribution at each time instant. This sampling process was then carried out forwards over the entire signal to generate a realisation.

\section{Experiments}
\label{sec:experiments}

The first case study considered in this section consists of an empirical simulation to assess whether the proposed method can recover the true data-generating system from a finite set of measurements and a finite number of MCMC samples. The second case study focuses on the identification of a nonlinear system using a benchmark dataset, namely the ``Silverbox'' dataset \cite{Wigren2013}. Finally, the third case study is experimental and investigates measured forces and velocities from the Christchurch Bay Tower \cite{Bishop1979}, an offshore structure designed to study fluid loading models in a real directional sea environment. 

\subsection{Academic case study and simulation}
\label{sec:academic_example}

The following academic example is replicated here from \cite{Khandelwal2020a}. The data-generating system used was a polynomial SISO model given by,
\begin{equation}
    y_i = 0.5u_{i-1} - 0.3u_{i-5}^3 + 0.3y_{i-1}
    \label{eq:academic_model}
\end{equation}

This structure delineates a NARX model. Although the model is fairly simple, the challenge in identifying this system is mostly provided by the cubic term in the equation. To simulate real measurements, Gaussian noise was added to the response. In particular,
\begin{equation}
    \hat{y}_i = y_i + \epsilon_i
    \label{eq:academic_data}
\end{equation}

The simulated datasets were then generated using a normally-distributed excitation signal $u_i \sim \mathcal{N}(0,1)$ and i.i.d. Gaussian noise $\epsilon_i \sim \mathcal{N}(0,\sigma_{\epsilon}^2)$. The noise variance $\sigma_{\epsilon}^2$ was adjusted such that the resulting signal had a \textit{signal-to-noise} ratio (SNR) of $20\mathrm{dB}$. The SNR is defined here as,
\begin{equation}
\mathrm{SNR}_{\mathrm{dB}} = 20 \log_{10} \left( \frac{\sigma_{y}}{\sigma_{\epsilon}} \right),
\label{eq:snr}
\end{equation}
where $\sigma_{y}$ denotes the standard deviation of the original (noise-free) response.
For the established scheme, the size of the generated training dataset was limited to $2000$ simulated samples. The first $1000$ samples were assigned as the training set, the subsequent $500$ as the validation set, and the remaining $500$ as the test set.

\subsubsection{Results and discussion}

The main results obtained from running the proposed algorithms are listed in Table \ref{tab:cs1}. These results include all the NMSEs computed on the held-out test set. The MAP and FWD notations denote whether the predictions were computed via MAP estimates or forward integration, respectively. Two sets of NMSE results are also listed in Table \ref{tab:cs1}; that is, ``selected'' or ``pooled''. The former indicates that the NMSE was computed from the chain that yielded the lowest error on the validation set, whereas the latter is the result from the union of all four chains. The truth column counts chains whose modal post-burn structure equals the true model, and the mass is the mean posterior probability assigned to that structure. Finally, the time column indicates, in hours per chain, the average time it took for the algorithm to run.
\begin{table}[htbp]
\centering
\caption{Results from Case Study 1. All scores are provided by the NMSE on the test set.}
\label{tab:cs1}
\begin{tabular}{llrrrrcrr}
\toprule
& & \multicolumn{2}{c}{Selected} & \multicolumn{2}{c}{Pooled} & & \multicolumn{2}{c}{Truth} \\
\cmidrule(lr){3-4}\cmidrule(lr){5-6}\cmidrule(lr){8-9}
Method & $\lambda_t$ & MAP & FWD & MAP & FWD & Time (h) & Rec. & Mass \\
\midrule
\multirow{3}{*}{OSA}
&  1 & 1.154 & 1.166 & 1.154 & \textbf{1.160} & 0.04 & 3/4 & 0.747 \\
&  5 & \textbf{1.154} & \textbf{1.166} & \textbf{1.154} & 1.165 & \textbf{0.03} & \textbf{4/4} & \textbf{0.987} \\
& 20 & 1.155 & 1.168 & 1.155 & 1.174 & 0.05 & 3/4 & 0.678 \\
\midrule
\multirow{3}{*}{MPO}
&  1 & 3.218 & 2.667 & 4.046 & 2.165 & 0.64 & 0/4 & 0.000 \\
&  5 & 1.154 & 1.164 & 3.173 & 1.748 & 0.71 & 1/4 & 0.254 \\
& 20 & 1.159 & 1.171 & 1.156 & 1.445 & 0.83 & 1/4 & 0.228 \\
\bottomrule
\end{tabular}
\end{table}

The OSA-based approach managed to discover the correct function on most occasions, even for different values of $\lambda_t$. The algorithm is also significantly more efficient computationally, completing the task at a much faster rate. The MPO-based algorithm managed to identify the correct function in two instances. In the failed cases, the trace of proposed models showed that the correct nonlinearity was indeed captured. The problem was that the sampler kept rejecting the $y_{i-1}$ term whenever it was proposed. This issue was likely caused by the way in which new parameters are proposed. Even when the structure may have been correct, a poor parameter-vector can make the fit worse. The OSA sampler mitigates this rejection since it draws a new set of parameters from the exact conjugate posterior at every move.

Table~\ref{tab:cs1params} lists the estimated parameter statistics that were taken over the post-burn draws from the validation-selected chain of each training objective, both of which landed at the true model structure. The intervals spanned by the estimated posteriors are shown as equal-tailed $95\%$ credible intervals ($2.5$--$97.5$ percentiles). All true parameter values were found within the estimated intervals. The main difference between the two approaches is that the bounds from the MPO-based approach are about $1.6$ times wider than those from the OSA-based approach. This outcome is to be expected since the ABC indicator gate discards information that the exact Gaussian likelihood retains, so the MPO posterior is less concentrated even when both samplers agree on the structure.
\begin{table}[htbp]
\centering
\small
\caption{Parameters identified on the validation-selected chain of each training objective, both at $\lambda_t=5$.}
\label{tab:cs1params}
\begin{tabular}{lrrcrc}
\toprule
& & \multicolumn{2}{c}{OSA}
  & \multicolumn{2}{c}{MPO} \\
\cmidrule(lr){3-4}\cmidrule(lr){5-6}
Term & True & Mean & $95\%$ CI & Mean & $95\%$ CI \\
\midrule
$u_{i-1}$      & $+0.500$ & $+0.5024$ & $[+0.4847,\, +0.5196]$ & $+0.5070$ & $[+0.4776,\, +0.5333]$ \\
$u_{i-5}^{3}$  & $-0.300$ & $-0.3011$ & $[-0.3062,\, -0.2958]$ & $-0.3023$ & $[-0.3111,\, -0.2930]$ \\
$y_{i-1}$      & $+0.300$ & $+0.2961$ & $[+0.2827,\, +0.3088]$ & $+0.2924$ & $[+0.2699,\, +0.3151]$ \\
\midrule
$\sigma_{\epsilon}$ & $0.1561$ & $0.1752$ & $[0.1682,\, 0.1829]$ & $0.1663$ & --- \\
\bottomrule
\end{tabular}
\end{table}

Unlike the term parameters, the $\sigma_{\epsilon}$ values are not comparable between the two approaches. In the case of the OSA-based approach, $\sigma_{\epsilon}$ is inferred and defined by an inverse-gamma posterior, whereas for the MPO-based approach, $\sigma_{\epsilon}$ has no posterior, and it is instead defined by the fixed proxy $\sqrt{\varepsilon_0/100}$, which is implied by the cold tolerance. Both approaches seem to have slightly overestimated the measured noise. Upon further inspection, it was found that this bias in the OSA sampler was contributed by setting the scale hyperparameter to $\beta_{\epsilon}=1.0$. In the case of the MPO-based approach, however, the cold tolerance may have needed one additional adaptive stage to converge more closely to the true posterior.

\begin{figure}[!h]
    \centering
    \includegraphics[width=\textwidth]{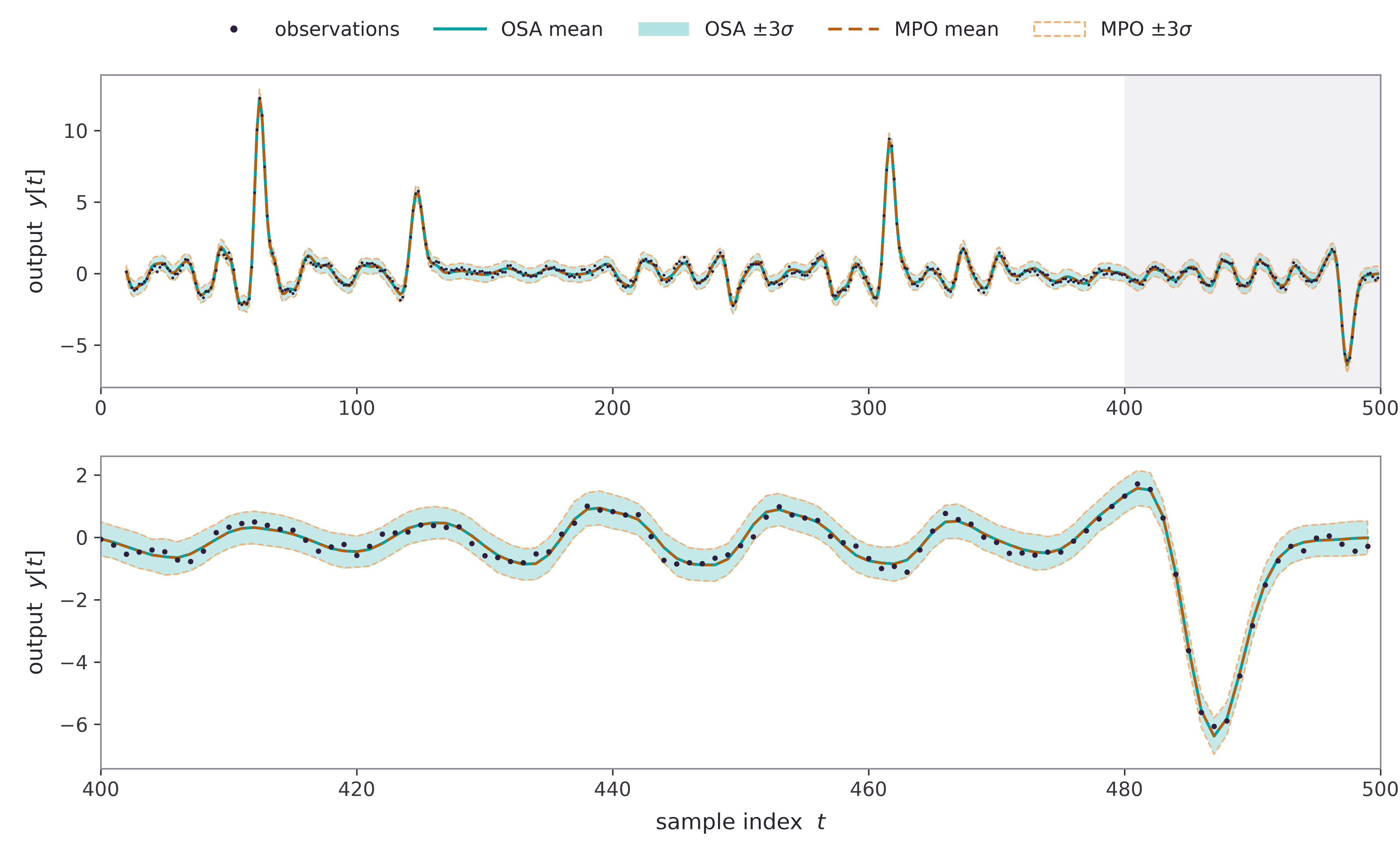}
    \caption{Comparison between response predictions and observations of the academic case study.}
    \label{fig:response_predictions}
\end{figure}

The forward integration procedure was repeated $100$ times to obtain Monte Carlo estimates of the response summary statistics. Figure \ref{fig:response_predictions} shows the posterior mean estimate and bounds corresponding to three standard deviations of the realisations.
The figure shows that selected chains fit the test data well. Additionally, the observations can be found within the estimated predictive bounds. While both approaches managed to correctly identify the data-generating system, the OSA-based approach clearly outperformed the MPO-based approach in terms of reliability and time consumption. This outcome may not apply to every scenario, as the next two case studies demonstrate.

\subsection{Nonlinear benchmarks: The Silverbox case study}

The Silverbox dataset is common among the datasets made available for benchmarking nonlinear system identifiers. The Silverbox system was proposed in Wigren and Schoukens \cite{Wigren2013} as an electronic implementation of the Duffing oscillator. The dataset is conveniently split into two parts for the identification problem. An excitation signal corresponding to a low-pass filtered Gaussian noise with a bandwidth of $200\mathrm{Hz}$ is used for the generation of the first $40{,}000$ samples. The excitation signal also presents a linearly increasing amplitude ranging from $0\mathrm{V}$ to $3\mathrm{V}$. Conversely, in the second part, the excitation signal consists of a stationary random odd multi-sine signal.

The dataset was partitioned here in the following manner: The first $80\%$ of the multi-sine signal was used as the training set $\mathcal{D}_{\mathrm{train}}$. The remaining $20\%$ of the multi-sine signal was used as a validation set $\mathcal{D}_{\mathrm{val}}$. The Gaussian excitation part of the dataset was used as the test set $\mathcal{D}_{\mathrm{test}}$, which is independent of $\mathcal{D}_{\mathrm{train}}$, and was used to estimate the simulation error once the inference procedure has identified the posteriors over the model candidates together with their associated parameters. A point to note is that the number of samples in this case study was dropped to $10{,}000/20{,}000$ because of the higher computational cost demanded from this large dataset. 

Based on the information provided about the system in Wigren and Schoukens \cite{Wigren2013}, the nonlinearity in the true system is said to be cubic in nature, and thus the p-NARX grammar $G_N$, can be considered to be a suitable model class for the current identification task. Except for the number of samples and the lag window, the hyperparameters selected in this case study remain the same as those introduced in Section \ref{sec:hyperparameter_selection}. The lag window was set to $\ell=50$, as this is the number of initial samples that is suggested to be treated as transient when reporting results on this dataset.

\subsubsection{Results and discussion}

The results obtained from the Silverbox data are listed in Table \ref{tab:cs2}. The notation in the table is consistent with the one specified in the previous case study. An additional column, however, has been included here listing the \textit{Root-Mean-Squared Error} (RMSE) derived from the selected chain's MAP NMSE. The RMSE is given in $\text{mV}$ for comparison with the published benchmark.
\begin{table}[htbp]
\centering
\small
\caption{Results from Case Study 2. All scores are provided by the NMSE on the test set. The number of terms here corresponds to the median across chains.}
\label{tab:cs2}
\begin{tabular}{llrrrrccc}
\toprule
& & \multicolumn{2}{c}{Selected} & \multicolumn{2}{c}{Pooled} & & & \\
\cmidrule(lr){3-4}\cmidrule(lr){5-6}
Method & $\lambda_t$ & MAP & FWD & MAP & FWD & No.\ Terms & Time (h) & RMSE (mV) \\
\midrule
\multirow{3}{*}{OSA}
 &  1 & 0.0061 & 0.0092 & 0.0061 & 0.0087 & 19.5 & 0.27 & 0.419 \\
 &  5 & \textbf{0.0055} & \textbf{0.0080} & 0.0200 & 0.0095 & 20.0 & 0.38 & \textbf{0.397} \\
 & 20 & 0.0068 & 0.0073 & \textbf{0.0065} & 0.0097 & 19.5 & 0.29 & 0.442 \\
\midrule
\multirow{3}{*}{MPO}
 &  1 & 7.2078 & 6.9726 & 8.6794 & 7.8068 & 3.0 & 7.50 & 14.358 \\
 &  5 & 0.7404 & 0.3286 & 7.1003 & 4.1935 & 3.5 & 7.66 & 4.602 \\
 & 20 & \textbf{0.0407} & \textbf{0.0504} & \textbf{0.0479} & \textbf{0.0956} & 4.5 & 8.11 & \textbf{1.079} \\
\bottomrule
\end{tabular}
\end{table}

The pooling errors provide some insights into the multimodality of the posterior space. Their departure from the selected MAP gives an indication of the disagreement among the chains. This consequence occurs because pooling takes the modal structure over the union of all post-burn traces, so it is not a simple average of the chains. For example, Table \ref{tab:cs2} shows that, at OSA $\lambda_t=5$, the pooled MAP is almost four times that of the selected chain, perhaps because the mode falls on a structure that is not favoured by all the chains. On the other hand, where the posterior is concentrated, both errors agree closely, as seen for OSA $\lambda_t=1$ and $20$.

The identified model structure that achieved the lowest error (OSA MAP $\lambda_t=5$) was,
\begin{equation}
    \begin{split}
        y_{i} =\ & -2.6143\, y_{i-1}^{3} + 2.4638\, y_{i-1}^{2} y_{i-2} - 1.8405\, y_{i-1} y_{i-2} y_{i-3} + 1.7409\, y_{i-1} + 1.6417\, y_{i-2} y_{i-3}^{2} \\
                & - 1.0511\, y_{i-2} - 0.5677\, y_{i-2}^{3} + 0.3683\, u_{i-1} + 0.2961\, u_{i-1} y_{i-1} y_{i-3} + 0.2410\, y_{i-4}^{3} \\
                & + 0.1676\, y_{i-5} - 0.1017\, u_{i-3} - 0.0906\, u_{i-4} - 0.0835\, u_{i-2} + 0.0455\, u_{i} \\
                & + 0.0175\, u_{i-5} - 0.0147\, y_{i-7} + 0.0068\, y_{i-1}^{2}- 0.0049\, y_{i-3}^{2} + 0.0031\, u_{i-7}
        \end{split}
    \label{eq:identified_silverbox}
\end{equation}
which was identified with a model mass of $0.354$. The parameters in \eqref{eq:identified_silverbox} correspond to the empirical MAP values estimated from the inferred posterior $p(\Theta_{\mathcal{T}}|\mathcal{T},\mathcal{D}_{\mathrm{train}})$. Unlike the previous case study, the true parameter values are unknown, so no guarantees can be made about the correctness of these parameters. Nevertheless, the predominant nonlinearity found in the identified model is cubic in nature, which is another characteristic that agrees with the Silverbox system as described in Wigren and Schoukens \cite{Wigren2013}. Indeed, one may observe that all nonlinear contributions are cubic in \eqref{eq:identified_silverbox}.

\begin{figure}[!h]
    \centering
    \includegraphics[width=\textwidth]{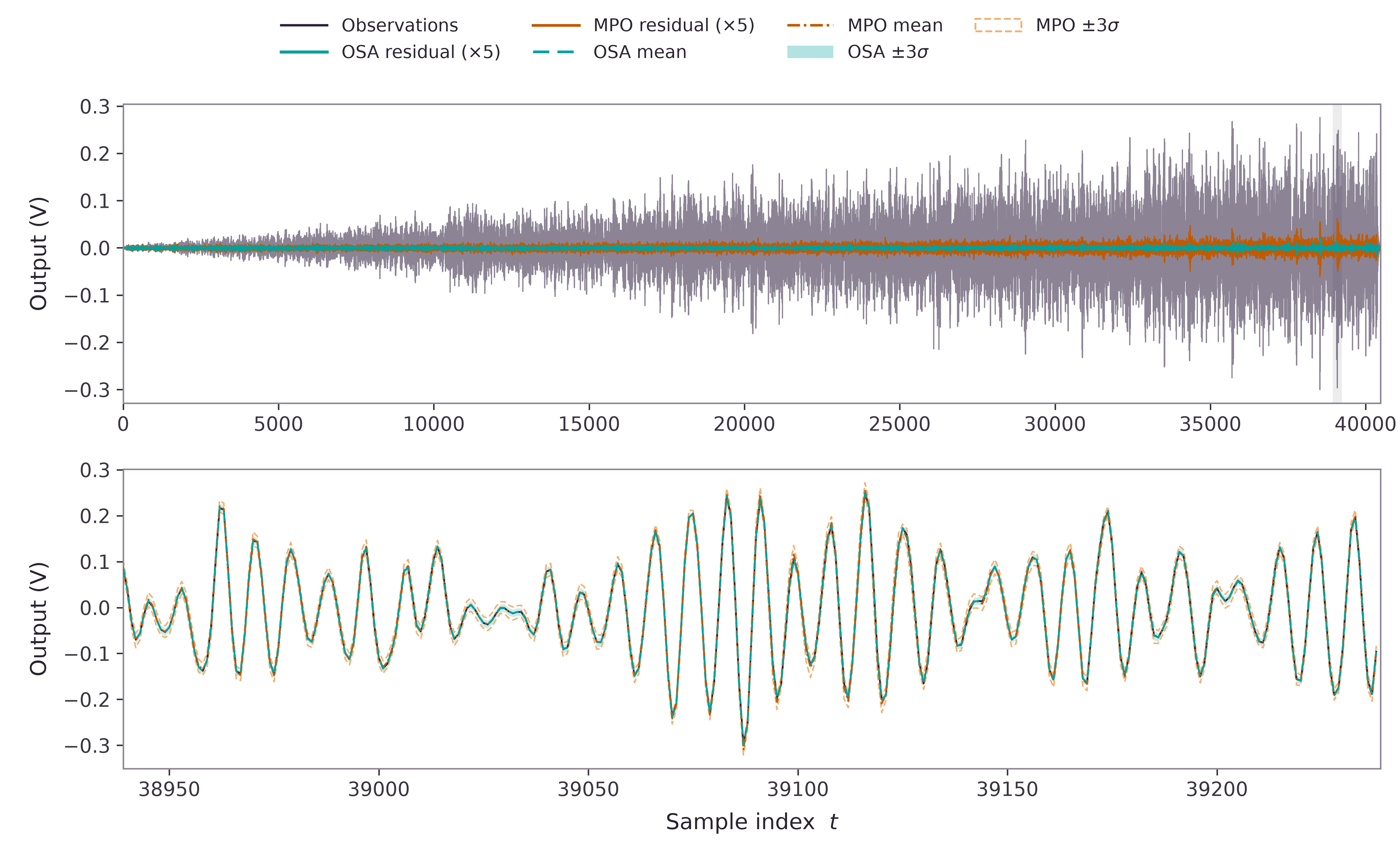}
    \caption{Comparison of the observations in the test set $\mathcal{D}_{\mathrm{test}}$ with predictions made by the validation-selected models.}
    \label{fig:response_predictions_silverbox}
\end{figure}

Figure \ref{fig:response_predictions_silverbox} (top) shows the residuals from the best MAP model-parameter pair plotted together with the test observations. Additionally, Figure \ref{fig:response_predictions_silverbox} (bottom) shows the estimated means and confidence bounds from $100$ realisations of the forward integration scheme. The zoomed-in view of the signal illustrates how well the models agree with the observations.

To compare these results against the published benchmark, Table \ref{tab:benchmark_silverbox} lists the results from other nonlinear identification methods proposed in the literature. The best result reported here remains on par with state-of-the-art methods.
\begin{table}[tp]
\caption{Performance of the identified model computed on the Silverbox test data and compared with other results in the literature.}
\label{tab:benchmark_silverbox}
\centering
\begin{tabular}{ccll}
\cline{1-2}
\multicolumn{1}{l}{Nonlinear identification method}         & \multicolumn{1}{l}{Test RMS simulation (mV)}  &  &  \\ \cline{1-2}
\textbf{Proposed method}                                    & \textbf{0.397}    &  &  \\
\cline{1-2}
Genetic Programming NARMAX (\cite{Khandelwal2020a})          & 0.360             &  &  \\
GP-NARX (\cite{Champneys2024})                              & 0.720             &  &  \\ 
p-NARX (\cite{Champneys2024})                               & 2.250             &  &  \\ 
\textbf{p-NLSS} (\cite{Paduart2010})                        & \textbf{0.260}    &  &  \\ \cline{1-2}
\end{tabular}
\end{table}

A few insights are worth highlighting when comparing the performance of these methods. The first is that the proposed approach performs generally better than the simple p-NARX implemented in \cite{Champneys2024}. This benchmark may serve as an upper bound on the acceptable prediction error that novel methods should improve upon to be worthy of consideration. The fairly large error observed from the p-NARX model derives from restricting the model to a weighted sum of monomials in the polynomial expansion. The inclusion of cross-term multinomials should naturally enhance identification, which is precisely the improvement observed in the present case study. 

Perhaps more notable is that the proposed method outperformed the \textit{Gaussian Process NARX} (GP-NARX) model, also implemented in \cite{Champneys2024}. This result carries important implications, since the combinatorial nature of the lag-structure selection problem can make the GP-NARX a particularly challenging model to cross-validate, despite its flexibility and probabilistic framework. This limitation is somewhat mitigated here because models are evaluated in a symbolic-regression fashion, which is computationally more efficient. In other words, the proposed method not only achieves an improved performance, but may do so at a reduced computational cost, while still preserving the Bayesian characteristics of the GP-NARX.

Finally, it is important to highlight the results presented by Khandelwal \textit{et al}.\ \cite{Khandelwal2020a}. Although both approaches rely on the same TAG representation of dynamical systems, the performance achieved with genetic programming remains slightly better. A possible explanation is that their employed grammar was richer, allowing it also to encode p-NARMAX models. By explicitly modelling the measurement noise, some additional system characteristics may have been captured that would otherwise remain unmodelled.

\subsection{Enhanced modelling with physics knowledge: The Christchurch Bay Tower case study}
\label{sec:christchurch_case_study}

The last case study presents the implementation of the proposed approach on a dataset collected from the project known as the Christchurch Bay Tower Compliant Cylinder (CBTCC) \cite{Najafian2000}. One of the main purposes of this tower was to test fluid-loading models in a real directional sea environment. The structure under investigation is composed of a large central column of $2.80\mathrm{m}$ in diameter and a small rigid cylinder of $0.480\mathrm{m}$ in diameter. The tower was equipped with an array of sensors, including perforated-ball velocity meters, pressure transducers, force sleeves, and wave buoys.

The data under consideration correspond to measurements collected exclusively from the smaller rigid cylinder. A region of $3000$ data points was selected where the flow was ensured to be mostly unidirectional. From this set, the data were split into three sequential subsets of $1000$ points for use as training, validation and test sets. 

\subsubsection{Towards a grey-box modelling approach using TAGs}

To achieve a high-fidelity wave-loading quantification, simulations from Computational Fluid Dynamics (CFD) are commonly elicited. The simulations rendered by CFD can be accurate and powerful, especially when modelling wave forces interacting with large and/or complex structures. While these high-fidelity analyses are explored in great detail within research communities, the high computational resources that come with them remain an ongoing issue. In industrial applications, it may be necessary to rely on more simplistic empirical models that provide approximated solutions at a reduced computational cost. An example of one such model is Morison's equation \cite{Morison1950}, which has been used extensively to model wave loading on slender members.

The original Morison's equation intentionally simplifies the hydrodynamic forces as the sum of a drag force and an inertial force. Given the kinematic wave velocity $u$, and acceleration $\dot{u}$, the force per unit axial length $F$ is given by,
\begin{equation}
    F = \frac{1}{2} \rho D C_d u |u| + \frac{1}{4} \pi \rho D^2 C_m \dot{u}
    \label{eq:morison_equation}
\end{equation}
where $\rho$ is the fluid density, $D$ is the cylinder diameter, $C_d$ is the drag coefficient and $C_m$ the inertia coefficient. Condensing the wave-loading forces down to two terms oversimplifies the physics involved in the wave-structure interaction. Consequently, Morison's equation struggles to model nonlinear phenomena, such as vortex shedding. The idea here is to demonstrate that a compromise between the high fidelity of CFD and the simplicity of Morison's equation is possible by employing a \textit{grey-box} \cite{Cross2022} modelling approach.

The term \textit{grey-box} derives from conceptually combining a \textit{white-box} with a \textit{black-box}. A model determined completely by physics is termed a white-box, while one determined purely by data is termed a black-box. The key idea here is to enhance the predictive capabilities of Morison's equation (white-box model) with the addition of a nonlinear p-NARX component (black-box model). Unlike the previous case studies, there are no compelling reasons to believe that the p-NARX model class can parsimoniously explain the data. In fact, it is generally unwise to construct a model driven uniquely by data when some of the physics are known. The current dataset is likely subject to various sources of unaccounted-for uncertainties, making system identification particularly challenging.

Enhancing black-box models with physics can be achieved in several ways. The simplest and most common approach is to form a direct sum of a white-box model with that of a black-box model. This approach involves fitting a white-box model first, followed by fitting a black-box on the residuals between the white-box predictions and observed data. The residuals may be interpreted as the physics that the white-box model fails to capture. A more flexible black-box model can then take on the task of identifying the remaining discrepancies. The predictions are hence computed as follows,
\begin{equation}
    y_i = F_i + f(u_i,\dot{u}_i) + \epsilon_i
    \label{eq:residual_learning}
\end{equation}

\begin{figure}[ht]
    \centering
    \includegraphics[width=\textwidth]{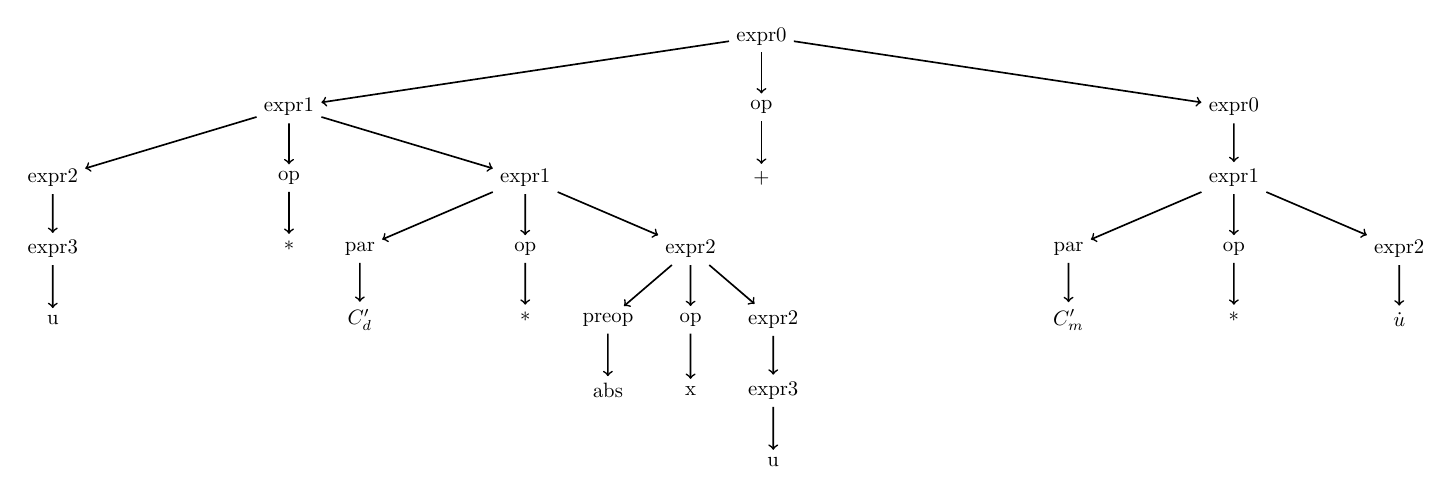}
    \caption{Morison's equation tree representation.}
    \label{fig:morison_equation_tree}
\end{figure}

The first component on the RHS corresponds to Morison's equation \eqref{eq:morison_equation}, and the second component to a p-NARX model encoded by a TAG. As stated in the introduction, the TAG framework is well suited to the incorporation of physics. This can be made possible by parsing Morison's equation into a valid tree representation, as shown in Figure \ref{fig:morison_equation_tree}. The yield of the tree is a version of Morison's equation in which the dimension-specific terms have been grouped to form the constants $C_{d}'$ and $C_{m}'$, relating to the drag and inertial forces of the wave, respectively. One may note that the tree encoding was extended to account for the ``abs'' (absolute value/norm operation) function as a pre-operation applied to the velocity factor $u$ within the first term.

The constructed tree representation in Figure \ref{fig:morison_equation_tree} can be enforced as the initial tree during the identification process. As a result, the model posterior remains conditioned on Morison's equation. This condition can be directly incorporated into the RJ-MCMC framework, since the jump kernel can be set to assign a zero probability to move types that could potentially modify the initial tree. By ensuring the initial tree remains fixed during its ramification, the yield is guaranteed to include the terms in Morison's equation within the overall expression.
The question then becomes how the white-box tree and its parameters should be treated during identification. Three possible alternatives are proposed here:
\begin{itemize}
    \item \textbf{Method 1 (Black-box):} Implement the proposed identifier directly using the TAG $G_N$. 
    \item \textbf{Method 2 (Residual):} Learn the parameters of the initial tree first. Then freeze the initial tree, along with its parameters, while allowing the remaining tree structure to grow around it during the identification process.
    \item \textbf{Method 3 (Grey-box TAG):} Set Morison's equation tree representation as the initial tree, and let the identifier grow the remaining tree over it while jointly learning all the parameters. 
\end{itemize}

Method $2$ ensures that the black-box model learns solely from the residuals. Meanwhile, Method $3$ relaxes this condition by letting the white-box parameters be somewhat corrected by the addition of new terms.

\subsubsection{Results and discussion}

\begin{figure}[ht]
    \centering
    \includegraphics[width=\textwidth]{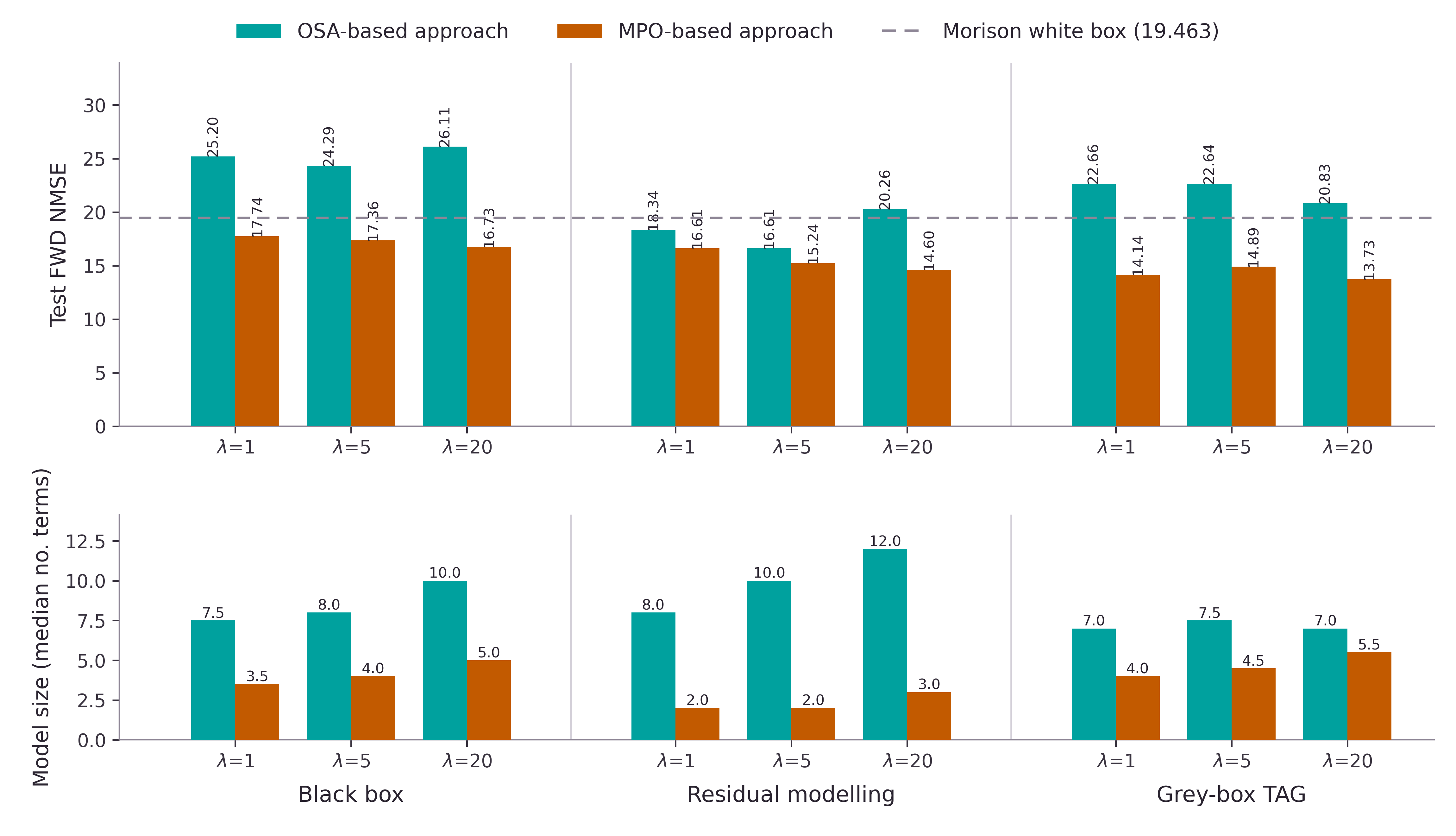}
    \caption{NMSE on the CBT test set and median number of terms for each method, training objective and $\lambda_t$, compared against the Morison's equation (white-box) baseline.}
    \label{fig:cs3_results}
\end{figure}

The results obtained in this case study are shown in Figure~\ref{fig:cs3_results}. In addition to the NMSE values, the figure also includes the model size that resulted from each run in terms of the median number of terms. All NMSE values are compared against the white-box baseline, which in this case yields an NMSE equal to $19.463\%$, as expected given that Morison's equation will typically have errors in the region of $20\%$ \cite{Wood1981}. This result was computed by fitting equation \eqref{eq:morison_equation} to the training data via least-squares regression.

In most cases, the MPO-based approach appeared to have outperformed the OSA-based approach. Furthermore, the latter infers model structures that are larger in every configuration. This outcome may suggest that OSA training on this dataset leads to overfitted models that heavily rely on true measured outputs. Only two cases of the OSA residual modelling managed to clear the Morison's white-box line. On the other hand, the MPO configuration manages to clear the line in all instances, with the best prediction $(\text{NMSE}=13.730\%)$ provided by the Grey-box MPO with $\lambda_t=20$. These results support the idea that a part of the structure found in the residuals could be captured by a more flexible polynomial NARX model.

\begin{figure}[ht]
    \centering
    \includegraphics[width=\textwidth]{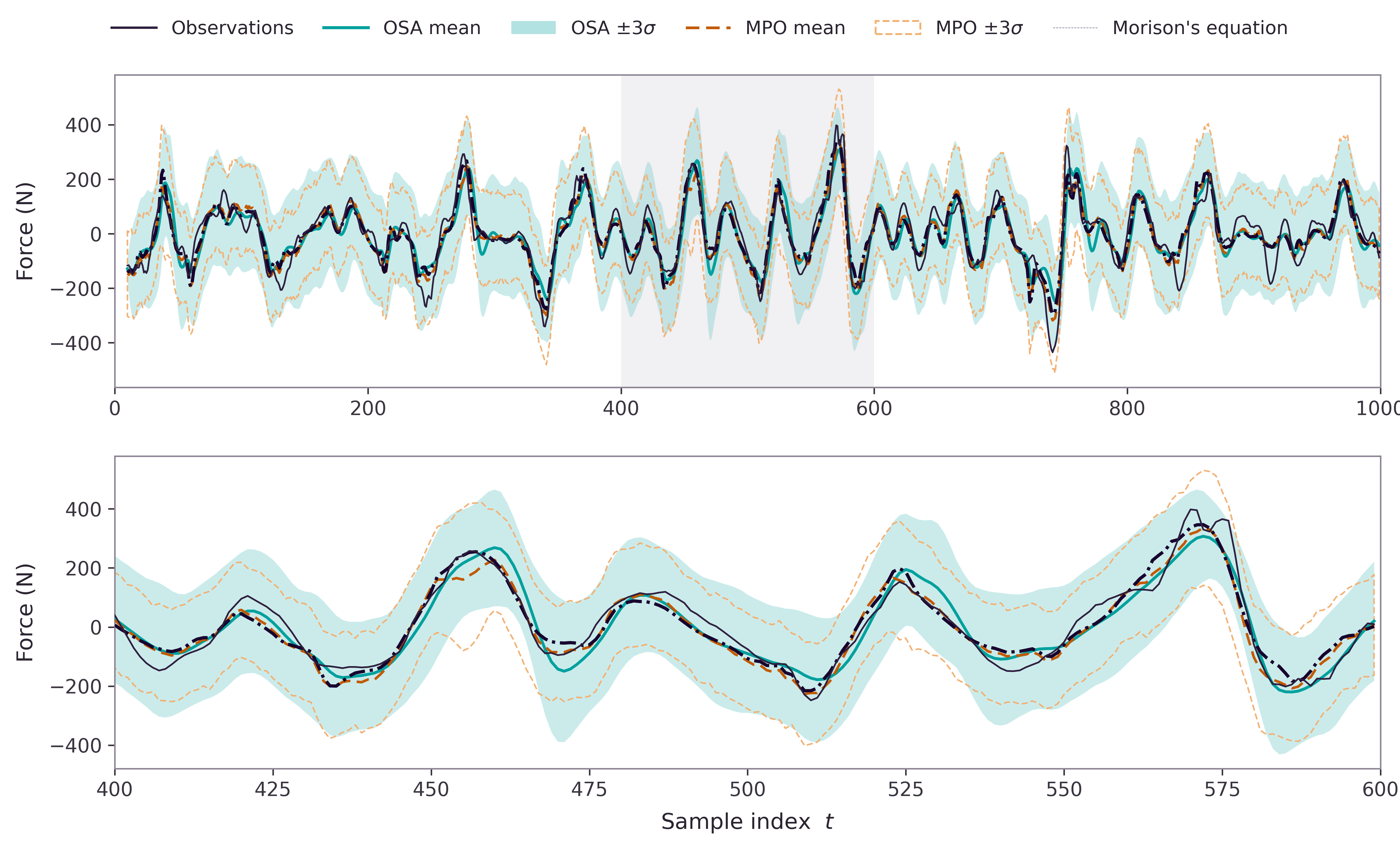}
    \caption{Comparison between observations of the CBT case study and response predictions.}
    \label{fig:cs3_test_prediction}
\end{figure}

The simulated predictions, alongside the test set observations, are shown in Figure \ref{fig:cs3_test_prediction}. Specifically, the figure shows the simulated force predictions from having identified the system using all methods. In this case, the Monte Carlo approximations were computed using $400$ realisations of the forward integration approach.
At first glance, and regardless of the modelling approach, the agreement of the predictions with the data seems to be fairly poor.
Nevertheless, the main wave-force trend can be seen to have been captured somewhat accurately, and the discrepancies are found within the estimated confidence bounds. To provide a better assessment of these results, Table~\ref{tab:cbt_nmse} lists the errors reported from other studies in which the CBT dataset has been modelled. In particular, the best results from \cite{Pitchforth2021,Worden2001,Worden2018} were included for comparison.

\begin{table}[tp]
\caption{Performance estimate comparison from identification methods used on the CBT test set. The proposed methods are reported for the MPO objective with $\lambda_t=20$.}
\label{tab:cbt_nmse}
\centering
\begin{tabular}{lcll}
\cline{1-2}
\multicolumn{1}{c}{Identification method}       & \multicolumn{1}{l}{NMSE ($\%$)}  &  &  \\ \cline{1-2}
\cline{1-2}
Morison's equation (white-box)              & $19.463$    &  &  \\ \cline{1-2}
Proposed method $1$ (black-box MPO)         & $16.730$    &  &  \\
Proposed method $2$ (residual MPO)          & $14.600$    &  &  \\
Proposed method $3$ (grey-box MPO)          & $\textbf{13.730}$    &  &  \\ \cline{1-2}
Polynomial NARMAX \cite{Worden2001}        & $21.430$    &  &  \\
GP-NARX \cite{Worden2018}        & $19.520$    &  &  \\
Morison's equation + GP-NARX \cite{Pitchforth2021}        & $13.840$    &  &  \\ \cline{1-2}
\end{tabular}
\end{table}

The relatively high errors reported in the literature demonstrate how hard it is to identify an autoregressive model that can accurately represent the CBT system. It becomes clear that confounded uncertainties contribute towards the ill-posed nature of this system-identification problem. One unaccounted source of uncertainty corresponds to the extent to which the flow is truly unidirectional. The available data-stream was cropped for a time window in which the load contribution from the orthogonal direction was at a minimum. While this is a sensible choice to ensure the effectiveness of Morison's equation, wave loads from all directions were never truly negligible. An additional source of uncertainty derives from the spatial offset between sensor equipment. The velocities were measured with the perforated-ball velocity meters attached at a distance of $1.228\mathrm{m}$ from the cylinder axis. Hence, these measurements do not provide the exact velocity at the centre of the force sleeve. Above all, however, the dataset may simply not be representative enough of the system. It is, perhaps, necessary to have a much larger dataset to discover an equation that can adequately reflect a complex system, like the CBT.

The grey-box model proposed by Pitchforth \textit{et al}.\ \cite{Pitchforth2021} managed to bring the NMSE down to $13.840\%$. In their original work, the same residual-modelling formulation as the one presented here was implemented, but with a GP-NARX regressor used instead as the black-box component. While the performance of the GP-NARX is comparable to that of Method $3$, the approach proposed in the present work offers several practical advantages that are similar to those highlighted in the Silverbox case study. The most prevalent benefit is that the model structure is autonomously determined by the Bayesian identifier. Conversely, to achieve the best performing GP-NARX for a given problem, an extensive validation scheme is necessary to determine which autoregressive terms, and their lags, should be used. Because of the high computational cost demanded by training a GP, and the complex combinatorial nature of NARX models, such a validation scheme may be too expensive to conduct within a reasonable amount of time.

\section{Overall discussion}
\label{sec:discussion}

The case studies examined above demonstrated the effectiveness of the proposed method in identifying nonlinear systems. An important point touched upon in all the case studies is the capability of the identifier to make predictions in a probabilistic framework. The uncertainty is not only quantified on the response predictions, but also estimated on both the model structure and associated parameters. Consequently, the model selection is no longer limited to a deterministic output that has to be assumed correct with certainty. A posterior mass function is rather estimated over a selection of model structures, each capable of explaining the data to some extent. Quantifying how certain one is about a model can be beneficial in informed decision-making and reliable extrapolation.

The previous section also demonstrated that the TAG can be easily augmented to include additional elementary trees. Such an extension was made in the CBT case study, not only by introducing an initial tree encoding the wave-loading physics, but also by introducing additional ``input'' trees, so that the TAG could parse \textit{Multi-Input-Single-Output} (MISO) autoregressive functions. Although made implicitly, the extension simply involved duplicating $\alpha_1$ and $\alpha_3$ in $G_N$ to account for both the velocity and acceleration signals. Even after making these changes to the grammar, the implementation of the sampler remained the same, with the small exception of the model prior \eqref{eq:distr_terms}, which required rewriting the factor $(1/2)^d$ as $(1/3)^d$, since the random choice was made uniformly among three input variables $\{u,\dot{u},y\}$.

Regarding the proposed approaches, deciding whether to implement the sampler based on OSA or MPO depends entirely on the application at hand. The OSA objective may be preferred to measure short-term predictions, while the MPO objective may be preferred for long-term predictions \cite{Khandelwal2020a}. While it has been argued that the MPO objective tends to be a more stringent test of the model \cite{Worden2001}, including both of these seemingly complementary measures is perhaps necessary. In the examined case studies, each approach displayed advantages over the other in different respects. For example, the OSA-based approach was easier to implement and significantly faster to compute. However, the model resulting from the MPO-based approach generalised better when the data-generating system did not belong to the model class. In fact, in such a scenario, the OSA-based approach fails to capture the dynamics of the underlying system. This difference was evident when comparing the results obtained from the Silverbox and CBT case studies.

A remark worth making on the CBT case study is that employing a polynomial expansion is, perhaps, an unsuitable choice to approximate the nonlinear nature of vortex shedding. While p-NARX models are considered universal approximators, the sampler may have needed to elicit a vast number of monomials to represent the true data-generating system, making the computational demand unfeasibly large. This issue was not experienced in the first two case studies because the nonlinearities in the systems are cubic in nature, and thus possible to identify within the p-NARX function class. Therefore, it may be necessary to explore the use of other TAGs, capable of representing distinct function classes that could be better suited for wave-loading problems. This research avenue is left for future work.

Finally, it is reiterated here that the chosen grammar, move types, priors, and hyperparameters are by no means the optimal selection. These configurations should instead be adjusted based on the knowledge one may have about the system being identified. Admittedly, having used the same priors across all three case studies may have defeated this purpose. This somewhat naive approach assumed that the contrasting datasets could be explained by the same underlying mechanisms. These assumptions concerned, for example, the expected level of measurement noise, or whether the model should promote pure-power terms over interaction terms. However, this choice was necessary to ensure the case studies remained as comparable as possible, albeit at the expense of performance on an individual basis.

\section{Conclusion}

Overall, the proposed algorithm addresses the nonlinear identification problem using TAGs from a Bayesian perspective. The presented case studies demonstrate that the Bayesian approach offers several advantages over existing methods, particularly via the seamless incorporation of prior information over model structures, and its ability to quantify uncertainty over the choice of suitable model structures, associated parameters, and predictions. This work may be regarded as a humble contribution to the broad range of algorithms developed for NLSI. Despite the promising results, the present study represents an early stage of the method, with the potential for further refinement and improvement.

\section*{Acknowledgments}
The authors of this paper gratefully acknowledge the support of the Engineering and Physical Sciences Research Council (EPSRC) Open Fellowship via Grant reference EP/X040852/1. The authors would also like to acknowledge Dr.\ Max Champneys for his insights on this work regarding Bayesian inference and nonlinear system identification. For the purpose of open access, the authors have applied a Creative Commons Attribution (CC BY) licence to any Author Accepted Manuscript version arising.

\bibliographystyle{unsrt}
\bibliography{references}

\appendix
\section{Algorithms}

\begin{algorithm}
    \caption{OSA-based Bayesian inference using TAGs for nonlinear system identification.}
    \label{alg:osa_main}
        \begin{algorithmic}[1]
            \Require Grammar $G_N$, prior parameters $\{\sigma_{\theta}^2,\alpha_{\epsilon},\beta_{\epsilon}\}$, reference tree $\eta_1$ (whose terms are held fixed), jump kernels $j(\mathcal{T} \to \mathcal{T}')$, and sample number $n_s$.
            \State Initialise $\mathcal{T} \gets \eta_1$ with one term adjoined, and $\sigma_{\epsilon}^2$ at the output scale.
            \State Sample $\Theta_{\mathcal{T}}$ from $\mathcal{N}(\Theta_{\mathcal{T}}|\mu_{\mathcal{T}},\Sigma_{\mathcal{T}})$.
            \For{$n_s$ number of samples}
            \State Propose move type $\mathcal{T} \to \mathcal{T}'$ with probability $j(\mathcal{T} \to \mathcal{T}')$.
            \If{move type is \textit{stay}}
                \State Retain the current structure, $\mathcal{T}' \gets \mathcal{T}$.
            \Else
                \State Modify current tree structure from $\mathcal{T}$ to $\mathcal{T}'$, according to chosen move type.
                \If{$\mathcal{T}'$ is unchanged, or any term of $\eta_1$ is absent from $\mathcal{T}'$}
                    \State Reject the proposal and continue to line 20.
                \EndIf
            \EndIf
            \State Sample $\Theta_{\mathcal{T}'}$ from $\mathcal{N}(\Theta_{\mathcal{T}'}|\mu_{\mathcal{T}'},\Sigma_{\mathcal{T}'})$, conditioned on the current $\sigma_{\epsilon}^2$.
            \State Evaluate acceptance probability given by \eqref{eq:rjmcmc_acc_prob_3}.
            \If{Proposed model and parameters are accepted}
                \State Update current state with proposals.
            \Else
                \State Remain in current state.
            \EndIf
            \State Sample new noise variance $\sigma_{\epsilon}^2$ according to the Gibbs update given by \eqref{eq:posterior_noise}.
            \EndFor
        \end{algorithmic}
\end{algorithm}

\begin{algorithm}
    \caption{MPO-based likelihood-free inference using TAGs for nonlinear system identification.}
    \label{alg:mpo_main}
        \begin{algorithmic}[1]
            \Require Grammar $G_N$, prior parameter $\{\sigma_{\theta}^2\}$, auxiliary proposal variance $\sigma_u^2$, reference tree $\eta_1$ (whose terms are held fixed), jump kernels $j(\mathcal{T} \to \mathcal{T}')$, sample number $n_s$, number of steps $C$, annealing stages $S$, quantile $q$, and distance $d(\mathcal{T},\Theta_{\mathcal{T}}) = \rho(\mathbf{y}^*,\mathbf{y})$ of the free-run simulation of $(\mathcal{T},\Theta_{\mathcal{T}})$.
            \Statex \textit{Ladder calibration}
            \State Draw random structures, apply a short greedy descent and polish their parameters; record the smallest distance reached as the floor $d_{\min}$.
            \State Place tolerances $\varepsilon_0 < \dots < \varepsilon_{C-1}$ geometrically in the excess $\varepsilon - d_{\min}$, with $\varepsilon_0$ just above $d_{\min}$.
            \State Initialise each step $c$ with a state satisfying $d(\mathcal{T},\Theta_{\mathcal{T}}) \le \varepsilon_c$.
            \Statex \textit{Sampling}
            \For{$n_s$ number of samples}
            \For{each step $c = 0,\dots,C-1$}
                \State Propose move type $\mathcal{T} \to \mathcal{T}'$ with probability $j(\mathcal{T} \to \mathcal{T}')$.
                \If{move type is \textit{stay}}
                    \State Retain the current structure, and perturb $\Theta_{\mathcal{T}'} \gets \Theta_{\mathcal{T}} + \mathcal{N}(0,s_c^2 \mathbb{I})$.
                \Else
                    \State Modify current tree structure from $\mathcal{T}$ to $\mathcal{T}'$, according to chosen move type.
                    \If{$\mathcal{T}'$ is unchanged, or any term of $\eta_1$ is absent from $\mathcal{T}'$}
                        \State Reject the proposal and continue to the next step.
                    \EndIf
                    \State Carry coefficients of retained terms across unchanged; draw $\mathbf{u}\sim\mathcal{N}(0,\sigma_u^2)$ for any term gained, and score under the same density any term lost.
                \EndIf
                \State Simulate the free run of $(\mathcal{T}',\Theta_{\mathcal{T}'})$ and evaluate the distance $d'$.
                \If{$d'$ is not finite or $d' > \varepsilon_c$}
                    \State Reject the proposal and continue to the next step.
                \EndIf
                \State Evaluate acceptance probability given by \eqref{eq:acceptance_ratio}.
                \If{Proposed model and parameters are accepted}
                    \State Update current state of step $c$ with proposals.
                \Else
                    \State Remain in current state.
                \EndIf
            \EndFor
            \State Every $n_{\mathrm{swap}}$ samples, for alternating adjacent pairs $(c,c+1)$: exchange states if $d_{c+1} \le \varepsilon_c$.
            \If{still annealing, and at the end of a stage}
                \State Update $d_{\min}$ to the smallest distance seen, tighten $\varepsilon_0$ onto the $q$-quantile of the cold step's distances, and re-space the remaining steps in the excess over $d_{\min}$.
                \State Reassign to each step a state satisfying its new tolerance.
            \EndIf
            \If{annealing has just finished}
                \State Freeze the ladder and the random-walk widths $s_c$.
            \EndIf
            \State After burn-in, record the state of the cold step $c=0$.
            \EndFor
        \end{algorithmic}
\end{algorithm}

\end{document}